%% file: main.tex
\documentclass[10pt,twocolumn,letterpaper]{article}

\usepackage{cvpr}              

\usepackage{microtype}
\usepackage{graphicx}
\usepackage{subcaption}
\usepackage{booktabs} 
\usepackage{multirow}
\usepackage{listings}
\usepackage{xurl}
\usepackage{enumitem}
\usepackage{float}
\usepackage[most]{tcolorbox}
\usepackage[table]{xcolor} 
\usepackage{amsmath}
\usepackage{amssymb}
\usepackage{mathtools}
\usepackage{amsthm}

\usepackage{marvosym}

\definecolor{bg_indep}{RGB}{235, 248, 249} 
\definecolor{bg_seq}{RGB}{254, 243, 235}   
\definecolor{bg_total}{RGB}{245, 245, 245} 

\lstdefinelanguage{json}{
    basicstyle=\ttfamily\footnotesize,
    numbers=left,
    numberstyle=\tiny,
    stepnumber=1,
    numbersep=8pt,
    showstringspaces=false,
    breaklines=true,
    frame=single,
    backgroundcolor=\color{gray!5},
    stringstyle=\color{blue},
    keywordstyle=\color{red},
    rulecolor=\color{black!40}
}
\lstdefinestyle{promptstyle}{
  basicstyle=\ttfamily\small,
  frame=single,
  breaklines=true,
  breakatwhitespace=false,
  columns=fullflexible,
  keepspaces=true,
  showstringspaces=false,
  showspaces=false,
  showtabs=false,
  tabsize=2,
  captionpos=b,
  xleftmargin=0.5em,
  xrightmargin=0.5em,
  backgroundcolor=\color{gray!5},
  rulecolor=\color{black!40}
}
\newtcolorbox{skillbox}{
  colback=gray!5,
  colframe=black!40,
  boxrule=0.4pt,
  arc=0pt,
  left=6pt,
  right=6pt,
  top=6pt,
  bottom=6pt,
  boxsep=0pt
}

\theoremstyle{plain}

\theoremstyle{definition}

\theoremstyle{remark}

\usepackage[textsize=tiny]{todonotes}

\usepackage[pagebackref,breaklinks,colorlinks,allcolors=cvprblue]{hyperref}
\usepackage[capitalize,noabbrev]{cleveref}

\definecolor{cvprblue}{rgb}{0.21,0.49,0.74}
\usepackage[pagebackref,breaklinks,colorlinks,allcolors=cvprblue]{hyperref}

\def\paperID{*****} 
\def\confName{CVPR}
\def\confYear{2026}

\title{ST-BiBench: Benchmarking Multi-Stream Multimodal Coordination in Bimanual Embodied Tasks for MLLMs}

\author{
    Xin Wu$^{1*}$, 
    Zhixuan Liang$^{2*}$,
    Yue Ma$^{3,4\textsuperscript{\Letter}}$,
    Mengkang Hu$^{2}$,
    Zhiyuan Qin$^{4}$,
    Xiu Li$^{1\textsuperscript{\Letter}}$ \\[2mm]
$^{1}$ Tsinghua University \quad
$^{2}$ The University of Hong Kong \\
$^{3}$ HKUST \quad
$^{4}$ Beijing Innovation Center of Humanoid Robotics \\
\url{https://bimanibench.github.io/}
}

\begin{document}
\maketitle
\input{sec/0_abstract}    
\input{sec/1_intro}

\input{sec/2_related_work}
\input{sec/3_problem_fomulation}
\input{sec/4_benchmark_build}
\input{sec/5_experiments}
\input{sec/6_conclusion}

{
    \small
    \bibliographystyle{ieeenat_fullname}
    \bibliography{main}
}
\nocite{*}
\input{sec/appendix}

\end{document}

%% file: sec/0_abstract.tex
\begin{abstract}
Multimodal Large Language Models (MLLMs) have significantly advanced the landscape of embodied AI, yet transitioning to synchronized bimanual coordination introduces formidable challenges in multi-stream multimodal integration. We introduce ST-BiBench, a comprehensive multi-tier framework for evaluating spatio-temporal multimodal coordination. Our approach centers on Strategic Coordination Planning, assessing high-level cross-modal reasoning over multiple action and perception streams. To investigate the "proximity paradox"—where semantically coherent plans fail to align with spatially grounded visual inputs—we incorporate Foundational Spatial Grounding to verify workspace awareness and arm-selection logic. Furthermore, we probe model frontiers through Fine-Grained Action Control, investigating whether MLLMs can directly synthesize high-dimensional continuous action modalities (16-Dim) from complex multimodal metadata. Evaluating 30+ state-of-the-art MLLMs, we uncover a persistent and pervasive "coordination paradox"—a significant gap between high-level strategic reasoning and fine-grained physical execution. Results reveal that while frontier MLLMs excel at logic-driven strategy, they frequently suffer from perception-logic disconnection and multi-stream interference during multimodal fusion. ST-BiBench provides a platform for identifying critical bottlenecks in multi-stream multimodal fusion and cross-modal alignment for complex embodied tasks.
\end{abstract}

%% file: sec/1_intro.tex
\section{Introduction}
\label{sec:intro}
\begin{figure*}[t] 
    \centering
    \vspace{4pt}
    \includegraphics[width=0.9\textwidth]{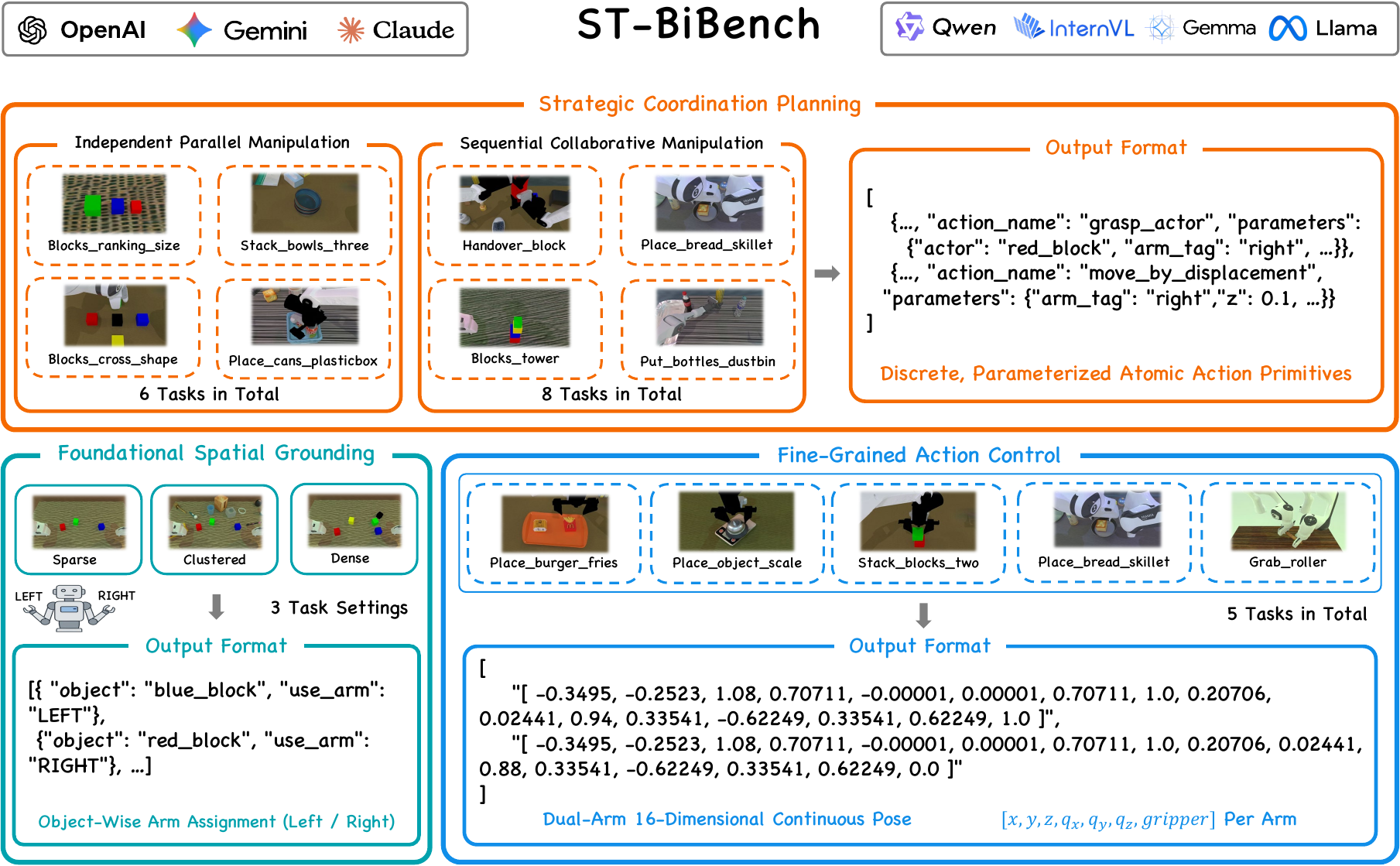} 
    \caption{\textbf{Overview of ST-BiBench.} An evaluation framework for MLLMs in bimanual manipulation, spanning three parts: Strategic Coordination Planning, Foundational Spatial Grounding and Fine-Grained Action Control.}
    \label{fig:framework}
    \vspace{-4pt}
\end{figure*}
Recent advances in Multimodal Large Language Models (MLLMs) have significantly enhanced the ability of artificial agents to interpret complex instructions and interact with physical environments. By integrating heterogeneous modalities—including vision, language, and action—these foundation models act as "brains" that bridge abstract language goals with grounded cross-modal reasoning~\cite{ahn2022can,driess2023palme,jiang2023vima}. Consequently, as MLLMs are increasingly deployed as direct controllers, evaluating their cognitive limits through comprehensive benchmarks has become a critical frontier in embodied intelligence within diverse and unstructured scenarios~\cite{huang2022language, yang2024embodied,o2024open}.

However, to accomplish increasingly sophisticated and realistic tasks, embodied agents must transition from single-arm manipulation to dual-arm (bimanual) coordination. Unlike single-arm operations, bimanual manipulation introduces the formidable challenge of multi-stream coordination across heterogeneous modalities, requiring synchronized reasoning over parallel perception-action channels. Bimanual setups significantly expand the operational workspace and enable complex, asymmetric interactions—such as handing over objects or stabilizing a container—that a single manipulator simply cannot achieve~\cite{fu2024mobile,zhao2023learning}. Yet, this transition introduces formidable cognitive and physical challenges, demanding rigorous spatio-temporal synchronization and the precise management of dual concurrent action streams. It requires the MLLM not only to recognize objects but also to understand the intricate inter-dependencies between two arms, such as preventing self-collisions and navigating overlapping kinematic workspaces. However, while existing embodied benchmarks (e.g., ALFWorld~\cite{shridhar2020alfworld}) have laid a solid foundation for assessing sequential reasoning within single-arm paradigms, a significant gap persists when addressing the evolving requirements of bimanual coordination. These dual-arm specific challenges—such as simultaneous multi-stream reasoning, dynamic role assignment, and mutual kinematic constraints—naturally fall outside the intended scope of traditional single-arm frameworks. As a result, the capacity for MLLMs to manage complex multi-stream multimodal fusion under concurrent action channels remains an open frontier. This necessitates a dedicated evaluation framework explicitly designed to assess MLLMs' proficiency in advanced, synchronized bimanual coordination scenarios, moving beyond the sequential constraints of prior single-manipulator studies.

To address this, we introduce \textbf{ST-BiBench}, a multi-granularity framework designed to evaluate the spatio-temporal multimodal coordination of MLLMs (Fig.~\ref{fig:teaser}). At the heart of our framework is Strategic Coordination Planning, which serves as the core reasoning engine for evaluating high-level cross-modal alignment over multiple action and perception streams. This module assesses the agent's ability to act as a strategic planner by decomposing long-horizon instructions into a sequence of atomic primitives~\cite{wang2023voyager}. These include \textit{independent parallel manipulation}, where two arms execute distinct sub-tasks simultaneously, and \textit{sequential collaborative manipulation}, where strict temporal dependencies exist. By focusing on high-level logic, we assess whether MLLMs can maintain "role consistency" and correct temporal sequencing over extended horizons.

During our evaluation, we observed a persistent "proximity paradox": models often generate logically sound plans but fail at the most basic level of optimal arm assignment. Unlike flexible humans, robotic manipulators operate within rigidly and strictly defined reachable workspaces. Forcing an arm to reach beyond its primary operational zone often triggers kinematic singularities and immediate execution collapse. This highlights a critical failure in aligning semantically coherent plans with spatially grounded visual inputs, a core challenge in multimodal learning~\cite{chen2024spatialvlm}. To systematically investigate this gap, we incorporate \textbf{Foundational Spatial Grounding} to verify the model's awareness of bilateral workspace constraints and its ability to resolve spatial-semantic conflicts. Furthermore, we explore the frontiers of MLLM capabilities through \textbf{Fine-Grained Action Control}. By providing rich context-augmented metadata, we test whether these models can directly synthesize high-dimensional continuous action modalities (16-Dim), investigating the limits of dense modality fusion from perception to fine-grained actuation in an end-to-end manner.

To support robust evaluation, we design a vision-driven agent framework featuring a Task-Adaptive Execution Truncation mechanism, balancing open-loop reasoning efficiency with closed-loop execution robustness in highly dynamic and complex environments.

Using ST-BiBench, we evaluate over 30 state-of-the-art MLLMs ~\cite{gpt5_openai,gpt4.1_openai,gpt4o_openai,achiam2023gpt,comanici2025gemini,gemini2.0flash,claude4,claude3.7,bai2025qwen3vltechnicalreport,Qwen2.5-VL,Qwen2-VL,Qwen-VL,wang2025internvl3.5,zhu2025internvl3,chen2024internvl,llama4_scout,llama3.2,team2025gemma,lu2025ovis25technicalreport,lu2024ovis,vteam2025glm45vglm41vthinkingversatilemultimodal}. Our empirical results reveal key insights into the current spatio-temporal reasoning gap:
(1) \textbf{Perception-Logic Disconnection:} Even models with superior high-level planning proficiency still exhibit varying degrees of stochastic spatial hallucinations, failing to align high-level strategy with physical grounding.
(2) \textbf{Disparity Between Strategy and Actuation:} Translating logical plans into precise continuous action modalities remains a formidable challenge, revealing a persistent performance bottleneck in fine-grained action synthesis.
(3) \textbf{Inefficient Multi-Stream Fusion:} While high-capacity models successfully leverage multi-perspective visual data, smaller models lack the reasoning bandwidth to integrate multiple visual streams and auxiliary modalities concurrently, where additional inputs often introduce distracting noise.

Our contributions are summarized as follows:
(1) We propose \textbf{ST-BiBench}, a multi-granularity framework centered on multi-stream embodied coordination for evaluating MLLMs.
(2) We establish a \textbf{multi-tier evaluative pipeline} for analyzing cross-modal alignment from perception to action, providing a holistic view of the agent's reasoning chain.
(3) We provide a comprehensive study of 30+ MLLMs, revealing a coordination paradox rooted in failures of multimodal fusion and alignment, and identifying the critical bottlenecks in dual-arm manipulation tasks.

%% file: sec/2_related_work.tex
\section{Related Work}
\subsection{MLLMs for Embodied Reasoning and Cross-Modal Alignment}
The evolution of Multimodal Large Language Models (MLLMs) has shifted Embodied AI from symbolic planning~\cite{huang2022language, yao2022react, singh2023progprompt} toward unified, end-to-end architectures~\cite{brohan2023rt2, liu2023llava, driess2023palme}. Central to this transition is multimodal fusion—the integration of heterogeneous modalities into a unified representation~\cite{baltrusaitis2018multimodal, atrey2010multimodal}. While MLLMs achieve impressive alignment in perception-centric tasks~\cite{radford2021clip, li2023blip, alayrac2022flamingo}, achieving consistent grounding in continuous action spaces remains an open challenge~\cite{zitkovich2023rt1, reed2022generalist}. Inspired by multi-stream architectures in video analysis~\cite{feichtenhofer2016convolutional, simonyan2014two}, recent research has begun interpreting coordination as a multi-stream fusion process. Unlike prior efforts relying on intermediate abstractions like code policies~\cite{liang2022code} or visual prompting~\cite{yang2023set, suris2023vipergpt}, we evaluate MLLMs as unified agents to investigate how multi-stream interference hinders cross-modal alignment under dynamic spatio-temporal constraints~\cite{zhang2025vlabench, shafiullah2022clip}.
\subsection{Benchmarks for Embodied and Multi-Stream Interaction}
Benchmarking is critical for systematically quantifying the limits of multimodal foundation models. Early environments such as \textit{ALFRED}\cite{shridhar2020alfred} and \textit{ALFWorld}\cite{shridhar2020alfworld} focused on high-level symbolic reasoning, often abstracting away physical execution. More recent and advanced benchmarks, including \textit{VLABench}\cite{zhang2025vlabench} and \textit{EmbodiedBench}\cite{yang2025embodiedbench}, evaluate perception-to-action pipelines but predominantly operate under single-stream or sequential interaction settings, where concurrent coordination is not explicitly modeled.

To address dual-arm complexities, platforms like \textbf{TWIN}\cite{grotz2025twin} and \textbf{RoboTwin}\cite{chen2025robotwin} provide high-fidelity simulations for policy training via reinforcement learning. Complementary to these efforts, ST-BiBench focuses on evaluating the zero-shot capacity of off-the-shelf MLLMs. Rather than training task-specific policies, we analyze how general-purpose models handle the multi-stream coordination and cross-modal alignment necessary for bimanual manipulation.
\subsection{Spatio-Temporal Challenges in Bimanual Coordination}
Bimanual manipulation necessitates precise spatial grounding and temporal synchronization~\cite{smith2012dual}, with sophisticated coordination patterns ranging from parallel execution to tightly coupled collaboration~\cite{grotz2025twin}. Prior research identifies that maintaining role consistency across multiple action streams is a formidable and primary hurdle in dual-arm systems. State-of-the-art MLLM-based agents, in particular, frequently exhibit unique failures such as spatial hallucinations and inconsistent role assignment over long horizons~\cite{liu2024survey, wang2023evaluation}. From a multimedia perspective, 
These failures represent breakdowns in complex multi-stream multimodal fusion, where inconsistencies across perception, reasoning, and action streams lead to cascading execution errors. Recent studies highlight a persistent "coordination gap" where strategic logic fails to manifest as precise physical output. By probing these misalignments across different levels of abstraction, our work provides a holistic view of the bottlenecks in current spatio-temporal reasoning for embodied systems.

%% file: sec/3_problem_fomulation.tex
\section{Problem Formulation}

We formalize the Coordination Paradox as the systemic disconnection between semantically coherent logic and spatially grounded execution. Similarly, the Proximity Paradox represents a failure in spatial-semantic mapping, where the model misjudges kinematic feasibility despite accurate target identification. And We employ two distinct scoring protocols tailored to the nature of the tasks, focusing on the model's ultimate ability to achieve goals and its awareness of physical workspace constraints.

\subsection{Success Rate for Manipulation Tasks}

For both Strategic Coordination Planning and Fine-Grained Action Control tasks, we report the standard Success Rate. A trial is considered successful only if the terminal environment state satisfies the programmatic verification of goal conditions (\textit{e.g.}, stable object-gripper contact, target region occupancy, or completion of a collaborative handover). 
Crucially, our benchmark defines success based on the \textbf{final state} rather than adherence to a single "gold standard" trajectory. Given the zero-shot nature of MLLMs, we recognize that multiple valid action sequences can lead to the same goal. By focusing on environment-driven feedback, we allow the models to leverage their diverse reasoning strategies—whether parallel or sequential—provided they ultimately resolve the task's intrinsic inter-dependencies without violating physical safety.

\subsection{Gaussian-Weighted Spatial Score for Grounding Tasks}

In the Foundational Spatial Grounding task, the model selects the appropriate arm (left or right) to interact with a target object, evaluating its ability to align visual spatial cues with action decisions. In tabletop bimanual settings, this choice is primarily governed by the \textbf{Proximity Principle} and \textbf{Kinematic Reachability}.

We define the arm closer to the object as optimal, as each manipulator operates within a limited reachable region. Cross-centerline reaching (i.e., selecting the distal arm) significantly degrades \textbf{manipulability}, as the arm approaches kinematic singularities where the Jacobian becomes ill-conditioned. This leads to reduced control precision and increased risk of inter-arm collisions in overlapping workspace regions. Therefore, proximal arm selection is essential for maintaining stable and safe manipulation.

To quantify this spatial awareness, we introduce a Gaussian-Weighted Spatial Score as a continuous evaluation metric. For each target object $i$, the corresponding score $S_i$ is computed based on its distance $d_i = |x_{pos}|$ from the workspace centerline:

\begin{equation}
\small 
S_i =
\begin{cases}
100, & \text{if } a_{\text{pred}} = a_{\text{gt}} \\
100 \cdot \exp\left(-d_i^2 / (2\sigma^2)\right), & \text{otherwise}
\end{cases}
\end{equation}

where $a_{\text{gt}}$ is strictly determined by the proximity of the target to each arm's primary reachability zone (sign of the $x$-coordinate). This formulation provides a soft penalty mechanism: 
(1) Errors near the boundary ($d_i \approx 0$) retain a high partial score, acknowledging that such cases are kinematically feasible for both arms and reflect inherent reasoning ambiguity. 
(2) Errors far from the center—where using the incorrect arm would lead to immediate physical failure—are penalized more heavily. This distance-based metric thus serves as a robust proxy for the model's underlying awareness of mutual kinematic constraints, moving beyond a simple visual classification task to quantitatively evaluate the physical grounding of the MLLM's complex spatial reasoning.

%% file: sec/4_benchmark_build.tex
\section{Benchmark Construction}
\label{sec:benchmark}
To systematically evaluate MLLMs’ ability in multi-stream embodied multimodal fusion, we introduce \textbf{ST-BiBench}, a framework that evaluates the spatio-temporal coordination chain across different granularities. Unlike prior benchmarks, ST-BiBench centers on the cross-modal alignment between high-level reasoning and low-level physical execution, probing the emergence of coordinated behavior without task-specific fine-tuning.

\subsection{Task Settings}
\medskip\noindent\textbf{Strategic Coordination Planning.}
This tier functionally serves as the foundational core reasoning engine, evaluating the agent’s sophisticated multi-stream temporal reasoning and decomposition capabilities. In this setting, the MLLM acts as a high-level strategic planner, autonomously synthesizing a logical sequence of parameterized atomic action primitives (e.g., \textit{Grasp}, \textit{Place}, \textit{Move}). We systematically assess whether models can maintain long-term temporal consistency and stable cross-modal role assignment under complex spatio-temporal inter-dependencies, where the generated strategic output must be strictly aligned with the synchronized operational requirements of dual concurrent action streams.

\medskip \noindent\textbf{Foundational Spatial Grounding.}
This foundational task evaluates the model's underlying ability to align visual spatial perception with discrete action decisions under stringent and inescapable physical constraints. Given a visual observation, the model must determine the optimal manipulator (left or right arm) to execute a grasp.
This tier verifies the agent's spatial common sense—a non-negotiable prerequisite for successful bimanual execution. Our preliminary analysis identified a persistent "proximity paradox": models often generate logically sound plans but fail to align semantic logic with spatial grounding, assigning the distal arm to a target. This reveals a critical misalignment between perception and reasoning, leading to physically infeasible actions and execution collapse.

\medskip \noindent\textbf{Fine-Grained Action Controlling.}
This tier explores the limits of multimodal fusion across heterogeneous representations, demanding the agent to directly synthesize high-dimensional continuous action modalities (16-Dim poses and gripper states).
Unlike traditional Vision-Language-Action (VLA) models, we evaluate zero-shot control emergence by providing rich, context-augmented metadata (e.g., precise coordinates and joint offsets). This setting reframes low-level control as an extreme case of dense modality fusion, investigating how a reasoning-heavy model integrates dense numerical signals with raw visual streams to synthesize precise, synchronized trajectories in an end-to-end manner.

We emphasize that the Fine-Grained Control tier is intentionally designed as a rigorous extreme density fusion stress test rather than a proposal for a realistic robotic controller. By providing context-augmented numerical metadata (16-Dim), we probe the MLLM's frontier in directly synthesizing high-dimensional continuous modalities from raw visual-textual tokens, revealing the current bottlenecks in high-fidelity cross-modal synthesis.

\subsection{Taxonomy of Multi-Stream Coordination}
We systematically categorize tasks into three modes based on the required level of spatio-temporal synchronization that must be maintained between parallel perception-action channels:

\medskip\noindent\textbf{Independent Parallel Manipulation.}
Two concurrent action streams execute distinct sub-tasks within a shared workspace, testing the model's underlying ability to resolve spatial interference and optimize arm allocation based on grounded perceptual proximity.

\medskip\noindent\textbf{Sequential Collaborative Manipulation.}
Tasks involve strict temporal dependencies, where one modality stream's output serves as a prerequisite for the other, testing the model's ability to maintain long-term logical coherence and cross-stream synchronization.

\medskip\noindent\textbf{Synchronous Collaborative Manipulation.}
Evaluated specifically in the Fine-grained Action Synthesis tier, this mode requires both concurrent streams to actuate simultaneously within the same timeframe, challenging the model’s high-frequency and fine-grained temporal alignment capabilities under dynamic conditions.

\begin{figure*}[t] 
    \centering
    \includegraphics[width=0.9\textwidth]{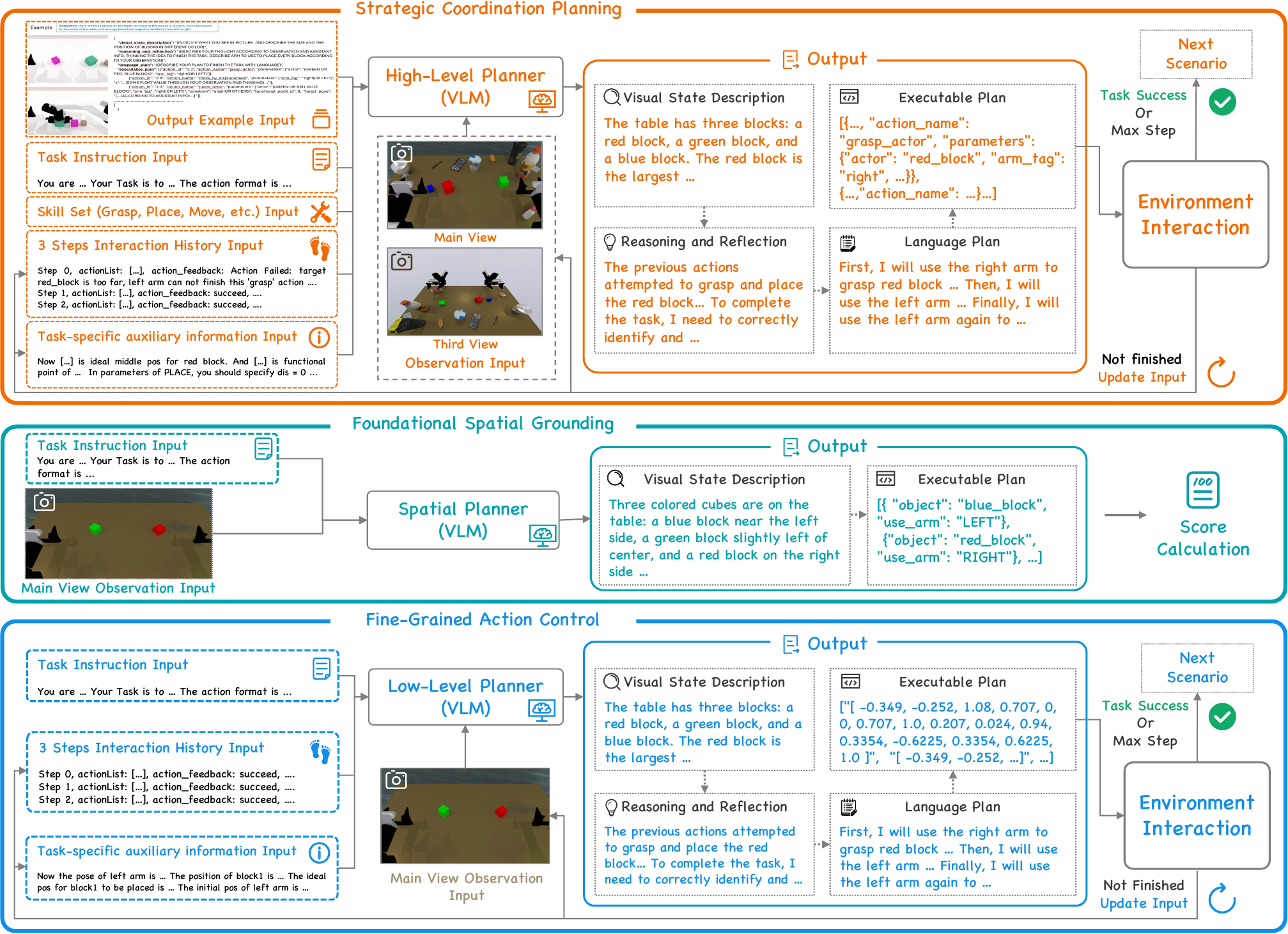} 
    \caption{Overview of the ST-BiBench agent architecture: A hierarchical framework for multi-stream input fusion and tiered action execution, maintaining consistent cross-modal alignment across perception, reasoning, and bimanual control.}
    \label{fig:vision-driven-agent}
\end{figure*}
\subsection{Vision-Driven Agent Design}
\label{agent_design}
To support the evaluation of complex multi-stream coordination, we design a unified vision-driven agent architecture that integrates diverse multimodal perception with reasoning-based planning through a synchronized processing pipeline as shown in Fig. ~\ref{fig:vision-driven-agent}.

\medskip\noindent\textbf{Heterogeneous Multi-stream Fusion.}
The agent dynamically operates on a  continuously synchronized stream of heterogeneous inputs to facilitate grounded reasoning: (1) Multi-perspective Visual Streams (ego-centric and third-person views) to mitigate occlusions and provide spatial redundancy; (2) Multimodal Metadata (numerical coordinates and offsets) to probe the limits of fine-grained control; and (3) Interaction Context (natural language and history). By fusing these concurrent modalities, the agent constructs a spatio-temporally unified internal representation necessary for resolving inter-arm dependencies in bimanual coordination.

\medskip\noindent\textbf{Reasoning-Driven Task Planner: Aligning Thinking and Action Streams.}
The agent employs a cross-modal "Describe-Reason-Plan" pipeline to ensure explicit alignment between the "Thinking Stream" (high-level semantic reasoning) and the "Action Stream" (low-level physical execution). To encourage temporal consistency, our framework supports Multi-step Action Chunking, allowing the agent to predict a sequence of consecutive actions in a single cycle. This modular cognitive design effectively reduces the risk of reasoning drift and maintains logical coherence across parallel perception-action channels during multi-stream coordination.

\medskip\noindent\textbf{Task-Adaptive Execution Truncation: Dynamic Recalibration.}
Effective bimanual manipulation is highly sensitive to the dynamic evolution of the multimodal environment. To mitigate "action lag" caused by rapid world-state changes and ensure real-time temporal alignment, we implement a Task-Adaptive Execution Truncation mechanism. This strategy balances long-horizon logical consistency with dynamic perceptual grounding by executing only the first k actions before re-observing the environment. This feedback loop ensures that the generated Action Stream remains strictly aligned with the evolving physical state, preventing execution collapse due to stale environmental data.

%% file: sec/5_experiments.tex
\section{Experiments}
\subsection{Experiment Settings}
\medskip\noindent\textbf{General Settings.}
To ensure robust statistical reliability and generalizability, we conduct a minimum of 100 independent evaluation episodes per individual task. We implement extensive and heterogeneous visual and physical domain randomization—systematically varying background textures, lighting conditions, and initial poses—to assess the robustness of multimodal fusion under perceptual shifts. This setup ensures that model performance reflects genuine multimodal generalization across non-stationary distributions rather than overfitting to specific environment seeds.

\medskip\noindent\textbf{Strategic Coordination Planning Settings.}
This primary component comprises a total of 14 diverse tasks categorized into \textit{Independent Parallel Manipulation} and \textit{Sequential Collaborative Manipulation}. The MLLM acts as a strategic engine, outputting a sequence of 7 parameterized action primitives (e.g., Grasp, Place, and Move) executed via an API-based pipeline. This abstraction isolates high-level coordination from low-level control noise, providing a direct measure of the model's internal "coordination brain" and its logical consistency across parallel perception-action channels.

\medskip\noindent\textbf{Foundational Spatial Grounding Settings.}
This tier specifically verifies the essential spatial awareness necessary for bimanual coordination, specifically investigating the "proximity paradox"—the sub-optimal assignment of distal arms to targets. We evaluate models across three configurations: (1) \textbf{Sparse}; (2) \textbf{Dense}; and (3) \textbf{Cluttered} (with distractors). This probes whether the MLLM's internal representation respects physical reachability limits and demonstrates accurate spatial-semantic alignment.

\medskip\noindent\textbf{Fine-Grained Action Control Settings.}
As an extreme stress test of dense modality fusion, this tier evaluates zero-shot control emergence via continuous 16-dimensional action streams. By providing context-augmented metadata (e.g., ground-truth poses) as auxiliary input, we investigate the MLLM's ability to fuse numerical metadata with raw visual streams to synthesize synchronized motor trajectories (7-Dim pose and 1-Dim gripper per arm), grounding discrete reasoning into continuous action modalities.

\subsection{Benchmark Results}
\medskip\noindent\textbf{Results on Strategic Coordination Planning}\\
Table \ref{tab:highlevel_results} presents the success rates for our core component, Strategic Coordination Planning. By analyzing the vertical performance gap across a diverse set of 30+ state-of-the-art MLLMs, we identify systemic patterns in how different model architectures handle the multi-stream coordination challenges.

\textbf{(1) Performance Ceiling and Reasoning Bandwidth.}
Proprietary models (Gemini-2.5-Pro, GPT-5) define the performance ceiling, maintaining significant success in high-complexity tasks like \textit{Place6} and \textit{Place7} where open-source counterparts collapse. This suggests that while all models face increased difficulty as coordination depth grows, proprietary giants possess a significantly higher "reasoning-level multimodal fusion" bandwidth. They exhibit superior robustness in maintaining cross-modal role consistency, whereas open-source models frequently suffer from perception-logic disconnection as the planning horizon extends.

\textbf{(2) Non-monotonic Scaling and Multi-stream Interference.}
Open-source models exhibit a compelling and observable non-monotonic scaling trend: mid-sized models (e.g., \textit{InternVL3-38B}, \textit{Qwen3-VL-32B}) frequently outperform their 70B+ or 200B+ counterparts. This "efficiency peak" indicates that for embodied tasks, architectural efficiency and instruction-following alignment are more critical than raw scale. Larger models appear more susceptible to multi-stream interference, where increased capacity leads to higher unintended stochasticity and hallucinations during long-horizon logic chains rather than improved precision.

\textbf{(3) The Reasoning Threshold and Multi-stream Collapse.}
Furthermore, a sharp and distinct "reasoning threshold" exists between the 7B and 30B+ tiers. Models in the 7B category (e.g., \textit{Qwen2.5-VL-7B}) exhibit a systemic and near-total collapse in collaborative tasks. For these smaller models, the introduction of a second action stream and complex temporal dependencies acts as perceptual noise, overwhelming their reasoning bandwidth and leading to immediate failures in role assignment. While the 30B+ tier enables basic bimanual logic, a significant "alignment gap" remains in tasks requiring strict cross-arm temporal synchronization.

\begin{table*}[t]
\centering
\caption{\textbf{Benchmark Results under the Strategic Coordination Planning.} We report the success rate (\%) for both independent parallel and sequential collaborative manipulation tasks. Detailed task descriptions are provided in supplementary materials S3.2. }
\label{tab:highlevel_results}

\setlength{\aboverulesep}{0pt}
\setlength{\belowrulesep}{0pt}
\renewcommand{\arraystretch}{1.2}

\resizebox{\textwidth}{!}{%

\begin{tabular}{l 
    >{\columncolor{bg_indep}}c cccccc  
    >{\columncolor{bg_seq}}c cccccccc  
    >{\columncolor{bg_total}}c         
}
\toprule
 & \multicolumn{7}{c}{\cellcolor{bg_indep}\textbf{Independent Parallel Manipulation Tasks}} & \multicolumn{9}{c}{\cellcolor{bg_seq}\textbf{Sequential Collaborative Manipulation Tasks}} & \\ 
\cmidrule(lr){2-8} \cmidrule(lr){9-17} 
\multirow{-2}{*}{\textbf{Models}} & \textbf{Avg.} & \cellcolor{bg_indep}Place1 & \cellcolor{bg_indep}Place2 & \cellcolor{bg_indep}Rank1 & \cellcolor{bg_indep}Rank2 & \cellcolor{bg_indep}Stack1 & \cellcolor{bg_indep}Stack2 & \textbf{Avg.} & \cellcolor{bg_seq}Hand1 & \cellcolor{bg_seq}Hand2 & \cellcolor{bg_seq}Hand3 & \cellcolor{bg_seq}Place3 & \cellcolor{bg_seq}Place4 & \cellcolor{bg_seq}Place5 & \cellcolor{bg_seq}Place6 & \cellcolor{bg_seq}Place7 & \multirow{-2}{*}{\textbf{Total Avg.}} \\ 
\midrule

\multicolumn{18}{c}{\textbf{Open-source Vision Language Models (70b+)}} \\
\midrule

Qwen3-VL-235B-A22B-Instruct~\cite{} & 58.67&36&	1&	75&	90&	83&	67&	50.88&	96&	63&	86&	92&	5&	17&	2&	46&54.21\\
Llama-4-Scout-17B-16E-Instruct$^{\dagger}$ &10.67 & 20 & 0 & 7 & 37 & 0 & 0 & 29.75 & 81 & 9 & 60 & 64 & 4 & 12 & 0 & 8 & 21.57\\
InternVL3-78B & 56.33&	72&	7&	69&	81&	39&	70&	33.63&	9&5	9&	26&	96&	4&	23&	0&	16&43.36\\
InternVL2.5-78B & 47.83 & 56 & 22 & 73 & 76 & 27 & 33 & 29.51 & 80 & 0 & 84 & 33 & 1 & 13 & 0 & 25 & 37.36\\
Qwen2.5-VL-72B-Instruct & 28.60 & 16 & 0 & / & 88 & 3 & 36 & 37.25 & 74 & 42 & 85 & 59 & 4 & 32 & 0 & 2 & 33.92\\
\midrule

\multicolumn{18}{c}{\textbf{Open-source Vision Language Models (27b+)}} \\
\midrule
InternVL3.5-38B & 41.50 & 72 & 0 & 6 & 81 & 52 & 38 & 33.13 & 94 & 11 & 18 & 93 & 12 & 8 & 0 & 29 & 36.71\\
InternVL3-38B & 57.50 & 71 & 0 & 79 & 77 & 58 & 60 & 49.38 & 63 & 67 & 90 & 97 & 1 & 44 & 0 & 33 & 52.86\\
InternVL2.5-38B & 45.33 & 1 & 0 & 83 & 70 & 62 & 56 & 33.00 & 92 & 11 & 84 & 3 & 7 & 47 & 0 & 20 & 38.29\\
Ovis2-34B & 45.50 & 80 & 2 & 77 & 82 & 1 & 31 & 31.75 & 96 & 3 & 14 & 94 & 10 & 27 & 0 & 10 & 37.64\\
Qwen3-VL-32B-Instruct & 54.67 & 41 & 16 & 88 & 75 & 40 & 68 & 50.88 & 93 & 63 & 75 & 96 & 14 & 36 & 8 & 22 & 52.50\\
Qwen2.5-VL-32B-Instruct & 52.67 & 62 & 7 & 93 & 88 & 24 & 42 & 50.13 & 94 & 55 & 88 & 95 & 7 & 49 & 0 & 13 & 51.21\\
Qwen3-VL-30B-A3B-Instruct & 19.83 & 7 & 0 & 61 & 23 & 15 & 13 & 26.25 & 51 & 1 & 54 & 51 & 5 & 44 & 0 & 4 & 23.50\\
Gemma-3-27b-it & 27.17 & 62 & 0 & 51 & 26 & 0 & 24 & 19.50 & 10 & 24 & 66 & 38 & 1 & 11 & 0 & 6 & 22.79\\
\midrule

\multicolumn{18}{c}{\textbf{Open-source Vision Language Models (7b+)}} \\
\midrule
Ovis2-16B & 27.50 & 67 & 0 & 25 & 32 & 38 & 3 & 24.88 & 72 & 0 & 1 & 97 & 4 & 25 & 0 & 0 & 26.00\\
Gemma-3-12b-it & 20.33 & 83 & 0 & 5 & 34 & 0 & 0 & 13.88 & 32 & 28 & 36 & 1 & 1 & 4 & 0 & 9 & 16.64\\
Llama-3.2-11B-Vision-Instruct & 6.50 & 1 & 0 & 23 & 15 & 0 & 0 & 20.63 & 68 & 5 & 8 & 69 & 1 & 14 & 0 & 0 & 14.57\\
Ovis2.5-9B & 17.83 & 41 & 0 & 47 & 16 & 0 & 3 & 28.75 & 78 & 14 & 71 & 33 & 7 & 27 & 0 & 0 & 24.07\\
InternVL3-8B & 13.83 & 55 & 0 & 16 & 3 & 6 & 3 & 10.38 & 0 & 1 & 71 & 0 & 0 & 11 & 0 & 0 & 11.86\\
InternVL2.5-8B & 2.67 & 2 & 0 & 13 & 1 & 0 & 0 & 1.25 & 0 & 9 & 0 & 0 & 1 & 0 & 0 & 0 & 1.86\\
Qwen2.5-VL-7B-Instruct & 1.67 & 3 & 0 & 4 & 2 & 1 & 0 & 1.25 & 1 & 0 & 6 & 0 & 3 & 0 & 0 & 0 & 1.43\\
\midrule

\multicolumn{18}{c}{\textbf{Closed-source Vision Language Models}} \\
\midrule
Gemini-2.5-Pro & 71.33 & 77 & 22 & 88 & 99 & 62 & 80 & 69.38 & 94 & 60 & 83 & 94 & 74 & 35 & 63 & 52 & 70.21\\
Gemini-2.5-flash & 67.17 & 60 & 36 & 82 & 93 & 64 & 68 & 59.00 & 78 & 28 & 78 & 94 & 61 & 35 & 45 & 53 & 62.50\\
Gemini-2.0-flash & 62.83 & 72 & 43 & 87 & 94 & 43 & 38 & 41.25 & 67 & 33 & 81 & 87 & 15 & 33 & 3 & 11 & 50.50\\
GPT-5 & 76.67 & 64 & 50 & 92 & 100 & 86 & 68 & 59.75 & 69 & 24 & 86 & 90 & 71 & 17 & 58 & 63 & 67.00\\
GPT-4.1 & 78.50 & 81 & 40 & 87 & 100 & 86 & 77 & 42.88 & 47 & 46 & 88 & 96 & 7 & 18 & 1 & 40 & 58.14\\
GPT-4o & 52.33 & 68 & 22 & 50 & 74 & 37 & 63 & 45.50 & 88 & 60 & 85 & 95 & 5 & 16 & 0 & 15 & 48.43\\
Claude-sonnet-4 & 67.00 & 57 & 31 & 68 & 97 & 82 & 67 & 46.63 & 92 & 32 & 83 & 97 & 37 & 20 & 1 & 11 & 55.36\\
Claude-sonnet-3.7 & 69.00 & 68 & 39 & 66 & 96 & 86 & 59 & 45.00 & 95 & 32 & 84 & 92 & 32 & 12 & 1 & 12 & 55.29\\
\bottomrule
\multicolumn{18}{l}{
\small
$^{\dagger}$ Llama-4-Scout-17B-16E-Instruct is a Mixture-of-Experts model with 109B total parameters and 17B activated parameters per forward pass.
}
\end{tabular}%
}

\end{table*}

\medskip\noindent\textbf{Results on Foundational Spatial Grounding}\\
To analyze the perceptual factors behind coordination failures in strategic planning, we evaluate spatial-semantic alignment across three scenario settings (Table~\ref{tab:spatial_results}). This tier examines whether the coordination paradox originates from a failure to map visual perception into physically feasible action decisions.

\textbf{(1) Closed-source Dominance and Scaling Trends.}
In alignment with observations from the strategic planning tier, closed-source models dominate, with Gemini and GPT families achieving over 90\% accuracy. Gemini-2.5-Pro (95.01\%) and Gemini-2.0-Flash (95.38\%) show particularly stable performance. Among open-source models, scaling trends are more evident than in planning: larger models generally improve spatial grounding. However, mid-sized models such as Ovis2-34B (92.67\%) and Qwen3-VL-32B (94.00\%) outperform several 70B+ models, indicating that vision-language alignment quality is more critical than parameter scale.

\textbf{(2) Environment Complexity: The Dynamics of Anchoring vs. Interference.}
Performance varies non-linearly from Sparse to Cluttered settings, revealing two opposing effects. 
\textit{Contextual Anchoring:} Larger models (e.g., Qwen3-VL-235B: 86.82\% → 93.50\%) benefit from denser scenes, where additional objects provide spatial references for more accurate relative positioning. 
\textit{Perceptual Interference:} Smaller models (e.g., Gemma-3-12B: 80.09\% → 57.17\%) degrade sharply under clutter, suggesting attention drift and failure to isolate target objects under multi-stream inputs.

\textbf{(3) Proximity Paradox.}
The results confirm that incorrect arm assignment remains a key bottleneck. Even high-performing models show drops under clutter (e.g., InternVL3-78B: 97.07\% → 90.16\%), indicating sensitivity to visual noise. These failures suggest that while models can infer \textit{what} to do, they often misalign spatial feasibility (\textit{which arm}) when multimodal inputs become complex, leading to execution failures in bimanual coordination.

\begin{table}[tb]
\centering
\caption{\textbf{Foundational Spatial Grounding Results.} We report the success scores across three scenario settings: Sparse, Dense, and Cluttered. "Avg." represents the overall mean performance.}
\label{tab:spatial_results}

\setlength{\aboverulesep}{0pt}
\setlength{\belowrulesep}{0pt}
\renewcommand{\arraystretch}{1.25} 
\setlength{\tabcolsep}{5pt} 

\resizebox{\columnwidth}{!}{
\begin{tabular}{l ccc >{\columncolor{bg_total}}c}
\toprule

 & \multicolumn{3}{c}{\textbf{Task scenario settings}} & \\

\cmidrule(lr){2-4}

\multirow{-2}{*}{\textbf{Models}} & 
\textbf{Sparse} & 
\textbf{Dense} & 
\textbf{Cluttered} & 
\multirow{-2}{*}{\textbf{Avg.}} \\
\midrule

\multicolumn{5}{c}{\textbf{Open-source Vision Language Models (70b+)}} \\
\midrule
Qwen3-VL-235B-A22B-Instruct    & 86.82 & 93.50 & 90.33 & 90.22 \\
Llama-4-Scout-17B-16E-Instruct & 85.49 & 87.75 & 86.16 & 86.47 \\
InternVL3-78B                  & 92.80 & 97.07 & 90.16 & 93.34 \\
InternVL2.5-78B                & 87.21 & 86.45 & 89.37 & 87.68 \\
\midrule

\multicolumn{5}{c}{\textbf{Open-source Vision Language Models (27b+)}} \\
\midrule
InternVL3-38B                  & 81.82 & 92.13 & 89.85 & 87.94 \\
InternVL3.5-38B                & 89.48 & 91.45 & 86.75 & 89.23 \\
InternVL2.5-38B                & 79.16 & 85.47 & 85.99 & 83.54 \\
Ovis2-34B                      & 94.78 & 92.78 & 90.45 & 92.67 \\
Qwen3-VL-32B-Instruct          & 94.47 & 95.77 & 91.77 & 94.00 \\
Qwen2.5-VL-32B-Instruct        & 85.16 & 86.08 & 87.38 & 86.21 \\
Qwen3-VL-30B-A3B-Instruct      & 85.50 & 91.13 & 88.98 & 88.54 \\
Gemma-3-27b-it                 & 92.40 & 81.12 & 85.78 & 86.43 \\
\midrule

\multicolumn{5}{c}{\textbf{Open-source Vision Language Models (7b+)}} \\
\midrule
Ovis2-16B                      & 94.07&	91.74&	88.00&	91.27 \\
Gemma-3-12b-it                 & 80.09&	57.17&	70.22&	69.16 \\
Llama-3.2-11B-Vision-Instruct  & 54.64&	53.62&	54.01&	54.09 \\
Ovis2.5-9B                     & 72.79 & 78.12 & 73.13 & 74.68 \\
InternVL3-8B                   & 79.53 & 69.79 & 86.79 & 78.70 \\
InternVL2.5-8B                 & 87.48&	78.96&	81.81&	82.75 \\
Qwen2.5-VL-7B-Instruct         & 75.20 & 65.83 & 79.34 & 73.46 \\
\midrule

\multicolumn{5}{c}{\textbf{Closed-source Vision Language Models}} \\
\midrule
Gemini-2.5-Pro                  & 96.14 & 96.77 & 92.12 & 95.01 \\
Gemini-2.5-flash                & 95.77 & 96.76 & 92.88 & 95.13 \\
Gemini-2.0-flash                & 95.45 & 98.69 & 92.00 & 95.38 \\
GPT-5                           & 94.73 & 95.13 & 92.97 & 94.28 \\
GPT-4.1                         & 93.43 & 92.48 & 91.76 & 92.55 \\
GPT-4o                          & 89.02	&91.13	&87.10	&89.08 \\
Claude-sonnet-4.5              & 96.12&94.78&92.23&	94.38 \\
Claude-sonnet-4                & 94.13 & 94.46 & 92.88 & 93.82 \\
Claude-sonnet-3.7              & 93.46 & 95.11 & 91.94 & 93.51 \\
GLM-4.5V                       & 91.48 & 97.77 & 93.00 & 94.08 \\

\bottomrule
\end{tabular}
}
\end{table}

\medskip\noindent\textbf{Results on Fine-Grained Action Control}\\
Finally, we evaluate the models' ability to manifest abstract logic into precise and synchronized motor trajectories (Table~\ref{tab:low_level_tasks}).

\textbf{(1) Gap in Precision Action Synthesis:}
A sharp and widening dichotomy exists between proprietary and open-source models. GPT-5 leads with 66.80\%, followed by Gemini-2.5-Pro at 60.20\%, while the top open-source model (InternVL3-78B) achieves only 27.60\%. This >30\% gap suggests that while open-source models have improved in high-level strategy, they still lack the fine-grained spatio-temporal understanding required for precise motor actuation. This highlights a persistent logic-action gap in the perception-to-actuation chain of current foundation models.

\textbf{(2) Disconnect Between General Intelligence and Embodiment:}
Notably, Claude-sonnet-4.5 scores 25.40\%, on par with the open-source Qwen3-VL-235B (25.20\%), despite its superior reputation in general reasoning tasks. This anomaly reinforces our hypothesis that general intelligence does not automatically translate to embodied proficiency. Furthermore, InternVL3-78B slightly outperforms the larger Qwen3-VL-235B, suggesting that model architecture and input resolution are more pivotal than parameter scale for synchronized, multi-stream action synthesis.

\begin{table}[tb]
\centering
\caption{\textbf{Fine-Grained Action Control Performance.} We report the \textbf{success rate (\%)} on specific atomic tasks. The rightmost column shows the average success rate. Detailed task descriptions are provided in supplementary materials S3.3. }
\label{tab:low_level_tasks}

\setlength{\aboverulesep}{0pt}
\setlength{\belowrulesep}{0pt}
\renewcommand{\arraystretch}{1.2} 
\setlength{\tabcolsep}{3pt}

\resizebox{\columnwidth}{!}{
\begin{tabular}{l ccccc >{\columncolor{bg_total}}c}
\toprule

 & \multicolumn{5}{c}{\textbf{Tasks}} & \\

\cmidrule(lr){2-6}

\multirow{-2}{*}{\textbf{Models}} & 
Place8 & Place9 & Place10 & Grab1 & Stack3 & 
\multirow{-2}{*}{\textbf{Avg.}} \\
\midrule

\multicolumn{7}{c}{\textbf{Open-source Vision Language Models}} \\
\midrule
Qwen3-VL-235B-A22B-Instruct    & 41 & 28 & 9  & 46 & 2  & 25.20 \\
Llama-4-Scout-17B-16E-Instruct & 1  & 0  & 0  & 29 & 0  & 6.00  \\
InternVL3-78B                  & 8 & 50 & 0 & 79 & 1 & 27.60 \\
Gemma-3-27b-it                 & 8  & 13 & 3  & 7  & 0  & 6.20  \\
\midrule

\multicolumn{7}{c}{\textbf{Closed-source Vision Language Models}} \\
\midrule
Claude-sonnet-4.5              & 17 & 13 & 6  & 89 & 2  & 25.40 \\
GPT-5                          & 66 & 83 & 50 & 79 & 56 & 66.80 \\
Gemini-2.5-Pro                 & 82 & 61 & 39 & 81 & 38 & 60.20 \\
Gemini-2.5-flash & 74 & 48 & 13 & 84 & 49 & 53.60\\
\bottomrule
\end{tabular}
}
\end{table}

\subsection{Impact of Multi-perspective Spatial Fusion}

State-of-the-art high-capacity models exhibit the potential to leverage auxiliary views for resolving visual ambiguity. For instance, Gemini-2.5-Pro improves significantly on \textit{Place9} (39\% → 55\%), suggesting a remarkable capacity for multi-perspective calibration and effective cross-view spatial reasoning in resolving complex occluded distances. Similarly, GPT-5 demonstrates a distinct gain on \textit{Grab1} (79\% → 84\%), indicating improved precision in grasping-point estimation when provided with redundant perceptual streams.

In contrast, less robust models frequently suffer from a significant "information overload" effect. InternVL3-78B and Gemma3-27B show consistent performance degradation, with Gemma3-27B dropping sharply on \textit{Place8} (13\% → 4\%). This divergence suggests that extra visual streams can act as multimodal noise rather than beneficial anchors, as these models frequently fail to align disparate ego-centric and third-person perspectives into a unified and coherent 3D representation. Instead of integration, the model's attention mechanism likely struggles to resolve the inherent perceptual redundancy, leading to cognitive confusion.

Even frontier architectures like GPT-5 exhibit minor fluctuations on \textit{Place8} and \textit{Place9} when the second view is introduced, indicating a subtle residual multi-stream interference where the model may prioritize less reliable visual cues over the primary stream. Overall, these results highlight that robust multi-view reasoning is a critical prerequisite for successful multimodal fusion in complex bimanual settings; simply increasing the number of visual inputs does not guarantee embodied proficiency without the ability to navigate the inherent conflict between different perspectives.

\begin{table}[t]
\centering
\caption{\textbf{Ablation on the impact of third-person view.} 
We report success rate (\%) as (Base / +Third View). }
\label{tab:ablation_third_view}

\setlength{\aboverulesep}{0pt}
\setlength{\belowrulesep}{0pt}
\renewcommand{\arraystretch}{1.2} 
\setlength{\tabcolsep}{8pt} 

\resizebox{0.8\linewidth}{!}{
\begin{tabular}{l ccc}
\toprule
 & \multicolumn{3}{c}{\textbf{Tasks}} \\
\cmidrule(lr){2-4}

\multirow{-2}{*}{\textbf{Models}} & \textbf{Place8} & \textbf{Grab1} & \textbf{Place9} \\
\midrule

\multicolumn{4}{c}{\textbf{Open-source Vision Language Models}} \\
\midrule
InternVL3-78B    & 50 / 43 & 79 / 73 & 0 / 1 \\
Gemma3-27B       & 13 / 4  & 7 / 6   & 3 / 0 \\
\midrule

\multicolumn{4}{c}{\textbf{Closed-source Vision Language Models}} \\
\midrule
Gemini-2.5-Pro   & 61 / 51 & 81 / 87 & 39 / 55 \\
Gemini-2.5-Flash & 48 / 47 & 84 / 87 & 13 / 12 \\
GPT-5            & 83 / 78 & 79 / 84 & 50 / 46 \\
\bottomrule
\end{tabular}}
\end{table}

\subsection{Error Analysis and Failure Diagnosis}
\begin{figure}[t]
    \centering
    \begin{subfigure}{0.45\columnwidth}
        \centering
        \includegraphics[width=\linewidth]{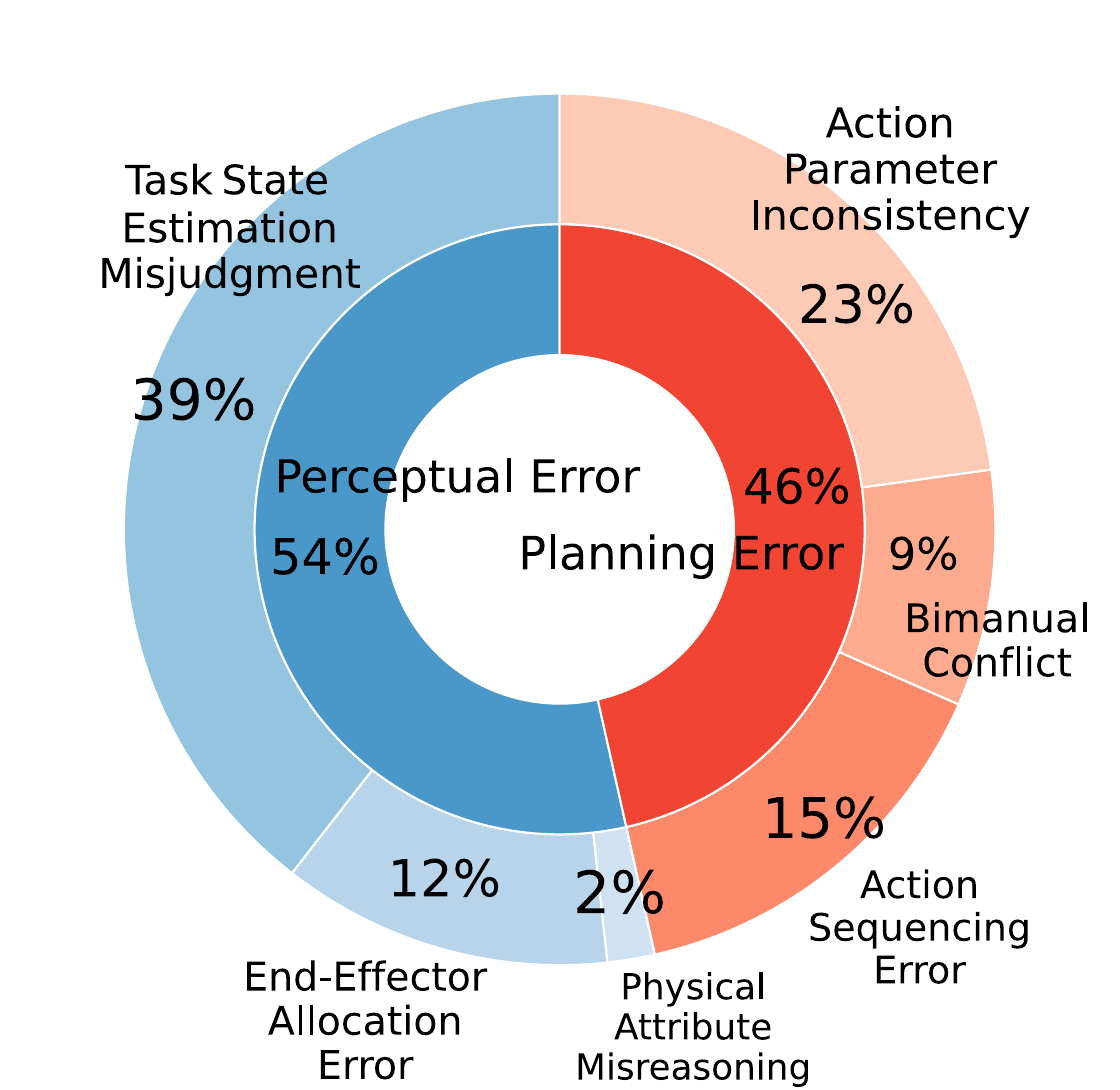}
        \caption{GPT-5}
        \label{fig:error_gpt5}
    \end{subfigure}
    \hfill
    \begin{subfigure}{0.45\columnwidth}
        \centering
        \includegraphics[width=\linewidth]{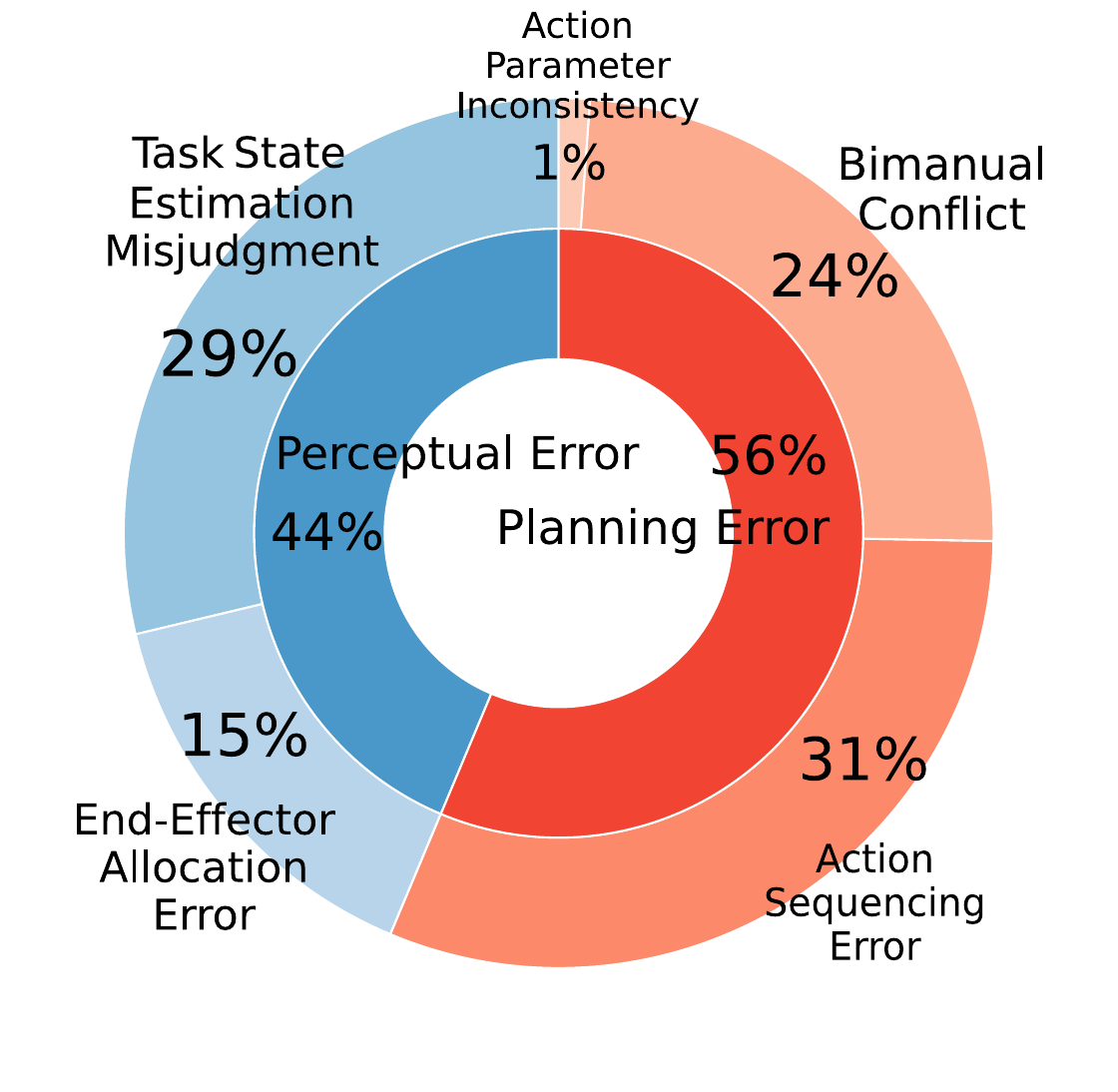}
        \caption{Gemini-2.5-Pro}
        \label{fig:error_gemini}
    \end{subfigure}
    \caption{Comparison of error type distributions.
     Analysis of failure modes for (a) GPT-5 and (b) Gemini-2.5-Pro. Inner rings represent error categories (Perceptual vs. Planning), while outer rings detail specific failure modes. Detailed definitions are provided in supplementary materials S4.}
    \label{fig:error_comparison}
\end{figure}

We analyze a granular and systematic failure patterns of GPT-5 and Gemini-2.5-Pro (Fig.~\ref{fig:error_comparison}) by grouping errors into high-level perceptual and planning categories, uncovering two starkly distinct limitations of current MLLM-based embodied agents.

\textbf{GPT-5: Perceptual Bottleneck.}
As shown in Fig.~\ref{fig:error_comparison}(a), GPT-5 is primarily constrained by perceptual misalignment (54\%). A major portion arises from \textit{Task State Estimation Misjudgment} (39\%), where incorrect scene understanding undermines otherwise valid plans, indicating an unstable internal world model. Additionally, \textit{Action Parameter Inconsistency} (23\%) highlights difficulty in grounding abstract plans into precise numerical actions. Although planning errors are lower (46\%), \textit{End-Effector Allocation} errors (12\%) persist, reflecting sensitivity to spatial grounding.

\textbf{Gemini-2.5-Pro: Planning Bottleneck.}
Conversely, Gemini-2.5-Pro exhibits a more reliable perceptual foundation (44\% error) but is significantly more limited by complex temporal planning logic (56\%), as illustrated in Fig.~\ref{fig:error_comparison}(b). Its main issues lie in \textit{Action Sequencing} (31\%) and \textit{Bimanual Conflict} (24\%), indicating difficulty in resolving temporal and spatial dependencies between dual action streams. While it adheres better to instructions, its reasoning degrades under concurrent coordination demands.

\textbf{Coordination Paradox.}
Together, these patterns highlight complementary weaknesses: GPT-5 fails in perception grounding, while Gemini-2.5-Pro struggles with multi-stream planning. Both exhibit notable \textit{End-Effector Allocation} errors (12\% vs. 15\%), confirming that arm selection remains a shared bottleneck. This suggests that bimanual coordination critically depends on consistent alignment across perception and planning; failure at either stage propagates through the entire perception–action loop.

%% file: sec/6_conclusion.tex
\section{Conclusion and Limitations}
We present \textbf{ST-BiBench}, a framework for evaluating MLLMs across the tiers of coordination planning, spatial grounding, and action control. Our results reveal that while frontier models demonstrate promising logic in \textbf{Strategic Coordination Planning}, this high-level reasoning capability varies significantly across the model landscape and remains brittle. Notably, \textbf{all evaluated models} exhibit pervasive \textbf{stochastic hallucinations} in \textbf{Foundational Spatial Grounding}, frequently misjudging arm proximity. These intermittent perceptual errors act as a critical bottleneck, disrupting otherwise sound strategic plans and leading to execution failures unique to dual-arm coordination. Furthermore, \textbf{Fine-Grained Action Control} remains a formidable challenge, where models struggle with multi-stream synchronization even when provided with extensive reasoning-supportive metadata.

ST-BiBench is currently simulation-based and confined to rigid-body tasks, which may not fully reflect the sensory nuances of real-world interaction. High inference latency also poses a hurdle for real-time coordination. While providing a robust end-to-end baseline, our framework does not yet support automated sub-goal failure analysis. We hope ST-BiBench identifies these persistent spatio-temporal gaps to guide the development of more physically-consistent and synchronized embodied agents.


%% file: sec/appendix.tex
\newpage
\appendix
\onecolumn

\setcounter{page}{1}

\setcounter{page}{1}
\renewcommand{\thepage}{S\arabic{page}}
\setcounter{section}{0}
\setcounter{figure}{0}
\setcounter{table}{0}

\section{Future Research Directions}
\label{app:future_research}
While ST-BiBench establishes a rigorous framework for evaluating MLLM-based agents in dual-arm coordination, it uncovers several persistent bottlenecks in multi-stream multimodal fusion. To catalyze further progress in bimanual embodied AI, we outline the following strategic research frontiers:

\begin{itemize}
    \item \textbf{Bridging the Multimodal Sim-to-Real Gap.}
    ST-BiBench currently operates in high-fidelity simulated environments (e.g., SAPIEN). However, real-world deployment introduces stochasticity such as sensor noise, non-uniform lighting, and mechanical latency. Future research should investigate cross-domain multimodal robustness, leveraging \emph{sim-to-real adaptation} and domain randomization to ensure that MLLM-derived coordination strategies remain resilient against the "perceptual drift" inherent in unstructured physical settings.

    \item \textbf{From Rigid Bodies to Multi-sensory Integration.} 
    The current benchmark focuses on rigid-body manipulation. Expanding the scope to deformable object manipulation (e.g., folding fabric, cable routing) presents a higher-dimensional challenge for spatio-temporal reasoning. Furthermore, incorporating tactile sensing streams as a primary modality alongside vision will be essential. Developing architectures capable of fusing high-frequency haptic feedback with visual streams is critical for achieving the delicate force-torque control required for complex bimanual interactions.
    
    \item \textbf{Architectural Innovations for Multi-stream Alignment.} 
    Our findings reveal that smaller MLLMs struggle with information overload and cross-stream interference. A pivotal direction is the design of specialized multi-stream attention mechanisms or hierarchical fusion modules. These architectures must selectively weigh inputs from ego-centric and third-person perspectives, resolving occlusions and spatial ambiguities without introducing distracting noise, thereby maximizing the "reasoning bandwidth" for bimanual tasks.
    
    \item \textbf{Temporal Synchronization and Latency Mitigation.} 
    High-precision manipulation requires high-frequency closed-loop control ($>50$~Hz), which remains a significant hurdle due to MLLM inference latency. Future work should explore hybrid control paradigms that bifurcate high-level strategic reasoning (via MLLMs) and low-level reactive execution (via high-speed policies). Techniques such as \emph{temporal ensemble action chunking} or model distillation will be vital to mitigate "action lag" and maintain real-time synchronization between the dual action streams.
    
    \item \textbf{Social-Physical Multi-stream Coordination.} 
    A natural evolution is extending ST-BiBench to human-robot bimanual collaboration. This requires the agent to manage not only its internal dual-arm constraints but also to perform intent recognition and safe physical interaction with a human partner. This transition toward multi-agent social coordination represents a new frontier for MLLMs, demanding more sophisticated cross-modal alignment across visual, linguistic, and social-physical feedback streams.

    \item \textbf{Predictive World Modeling via Video Generation.} 
    Current MLLM-based agents primarily operate in a reactive manner, selecting actions based on instantaneous observations. However, effective bimanual manipulation often requires foresight into long-horizon dynamics. A promising direction is to incorporate video generation~\cite{huang2025photodoodle, song2025omniconsistency, song2024diffsim, chen2025taming, chen2025s2guidancestochasticselfguidance, ma2024followpose, ma2024followyouremoji, ma2025controllable, ma2025followfaster, ma2025followyourmotion, ma2025followcreation} as a form of predictive world modeling, where agents synthesize future visual trajectories conditioned on candidate actions. Such imagination-based simulation enables agents to evaluate multiple strategies before execution, improving planning robustness, safety, and sample efficiency. Integrating action-conditioned video generation with MLLMs may thus provide a scalable mechanism for temporal reasoning and decision-making in complex embodied environments.
\end{itemize}

By addressing these directions, the community can bridge the current gap between high-level strategic planning and precise physical actuation, ultimately unlocking the full potential of bimanual embodied agents.

\section{Model Details}
\label{app:model_details}

In this section, we provide a detailed overview of the Multimodal Large Language Models (MLLMs) employed in our evaluation. We selected a diverse set of models ranging from state-of-the-art proprietary systems to open-source models of various parameter scales.

\paragraph{GPT, Gemini, and Claude Closed-Source Model Series}
For proprietary closed-source models, we assess the leading systems from OpenAI, Google, and Anthropic. The GPT series evaluation includes the flagship \textbf{GPT-5}, \textbf{GPT-4.1}, and the efficient \textbf{GPT-4o} ~\cite{gpt5_openai,gpt4.1_openai,gpt4o_openai,achiam2023gpt}. From Google, we test the Gemini family, specifically \textbf{Gemini-2.5-Pro}, \textbf{Gemini-2.5-flash}, and \textbf{Gemini-2.0-flash} ~\cite{comanici2025gemini,gemini2.0flash}. The Anthropic lineup features the Sonnet variants: \textbf{Claude-sonnet-4.5}, \textbf{Claude-sonnet-4}, and \textbf{Claude-sonnet-3.7} ~\cite{claude4,claude3.7}.

\paragraph{Qwen-VL Series} 
We evaluate multiple iterations of the Qwen-VL family, covering both the Qwen3 and Qwen2.5 generations. Specifically, we test the large-scale \textbf{Qwen3-VL-235B-A22B-Instruct}, along with the medium-sized \textbf{Qwen3-VL-32B-Instruct} and the mixture-of-experts (MoE) based \textbf{Qwen3-VL-30B-A3B-Instruct}. From the previous generation, we include \textbf{Qwen2.5-VL-72B-Instruct}, \textbf{Qwen2.5-VL-32B-Instruct}, and the lightweight \textbf{Qwen2.5-VL-7B-Instruct} ~\cite{bai2025qwen3vltechnicalreport,Qwen2.5-VL,Qwen2-VL,Qwen-VL}.

\paragraph{InternVL Series}
The InternVL family represents a significant portion of our open-source benchmarks. Our evaluation includes the latest \textbf{InternVL3} series (78B, 38B, and 8B variants) and the \textbf{InternVL3.5-38B} model. We also compare these against the widely used \textbf{InternVL2.5} lineup, encompassing the 78B, 38B, and 8B versions to analyze performance across different model capacities ~\cite{wang2025internvl3.5,zhu2025internvl3,chen2024internvl}.

\paragraph{Llama, Gemma, Ovis and GLM Series}
We also include other prominent open-weight models. The Llama series is represented by the advanced \textbf{Llama-4-Scout-17B-16E-Instruct} and the lightweight \textbf{Llama-3.2-11B-Vision-Instruct} ~\cite{llama4_scout,llama3.2}. Google's open models are tested using \textbf{Gemma-3-27b-it} and \textbf{Gemma-3-12b-it} ~\cite{team2025gemma} Additionally, we evaluate the Ovis family, including \textbf{Ovis2-34B}, \textbf{Ovis2-16B}, and \textbf{Ovis2.5-9B}~\cite{lu2025ovis25technicalreport,lu2024ovis}. And we also include \textbf{GLM-4.5V} as a representative model from the GLM family to ensure a comprehensive coverage of current high-performance MLLMs ~\cite{vteam2025glm45vglm41vthinkingversatilemultimodal}.

\begin{table}[H]
\centering
\caption{Full names of MLLMs used in our experiments.}
\label{tab:model_fullnames}
\resizebox{0.8\textwidth}{!}{
\begin{tabular}{c|cc}
\toprule
Model Name & Creator & Full Name \\
\midrule
GPT-5 & OpenAI & gpt-5-2025-08-07 \\
GPT-4.1 & OpenAI & gpt-4.1-2025-04-14 \\
GPT-4o & OpenAI & gpt-4o \\
Gemini-2.5-Pro & Google & gemini-2.5-Pro \\
Gemini-2.5-flash & Google & gemini-2.5-flash \\
Gemini-2.0-flash & Google & gemini-2.0-flash \\
Claude-sonnet-4.5 & Anthropic & claude-sonnet-4-5-20250929 \\
Claude-sonnet-4 & Anthropic & claude-sonnet-4-20250514\\
Claude-sonnet-3.7 & Anthropic & claude-3-7-sonnet-20250219 \\
\midrule
Qwen3-VL-235B-A22B-Instruct & Qwen & Qwen/Qwen3-VL-235B-A22B-Instruct \\
Qwen3-VL-32B-Instruct & Qwen & Qwen/Qwen3-VL-32B-Instruct \\
Qwen3-VL-30B-A3B-Instruct & Qwen & Qwen/Qwen3-VL-30B-A3B-Instruct \\
Qwen2.5-VL-72B-Instruct & Qwen & Qwen/Qwen2.5-VL-72B-Instruct \\
Qwen2.5-VL-32B-Instruct & Qwen & Qwen/Qwen2.5-VL-32B-Instruct \\
Qwen2.5-VL-7B-Instruct & Qwen & Qwen/Qwen2.5-VL-7B-Instruct \\
\midrule
InternVL3.5-38B & OpenGVLab & OpenGVLab/InternVL3\_5-38B \\
InternVL3-78B & OpenGVLab & OpenGVLab/InternVL3-78B \\
InternVL3-38B & OpenGVLab & OpenGVLab/InternVL3-38B \\
InternVL3-8B & OpenGVLab & OpenGVLab/InternVL3-8B \\
InternVL2.5-78B & OpenGVLab & OpenGVLab/InternVL2\_5-78B \\
InternVL2.5-38B & OpenGVLab & OpenGVLab/InternVL2\_5-38B \\
InternVL2.5-8B & OpenGVLab & OpenGVLab/InternVL2\_5-8B \\
\midrule
Llama-4-Scout-17B & Meta & meta-llama/Llama-4-Scout-17B-16E-Instruct \\
Llama-3.2-11B-Vision-Instruct & Meta & meta-llama/Llama-3.2-11B-Vision-Instruct \\
\midrule
Gemma-3-27b-it & Google & google/gemma-3-27b-it \\
Gemma-3-12b-it & Google & google/gemma-3-12b-it \\
\midrule
Ovis2-34B & AIDC-AI & AIDC-AI/Ovis2-34B \\
Ovis2-16B & AIDC-AI & AIDC-AI/Ovis2-16B \\
Ovis2.5-9B & AIDC-AI & AIDC-AI/Ovis2.5-9B \\
\midrule
GLM-4.5V & ZhipuAI & glm-4.5v \\
\bottomrule
\end{tabular}%
}
\end{table}

\clearpage
\section{Task Details}
\subsection{Foundational Spatial Grounding}

\medskip\noindent\textbf{Examples}

\begin{figure}[H]
    \centering
    \includegraphics[width=0.8\textwidth]{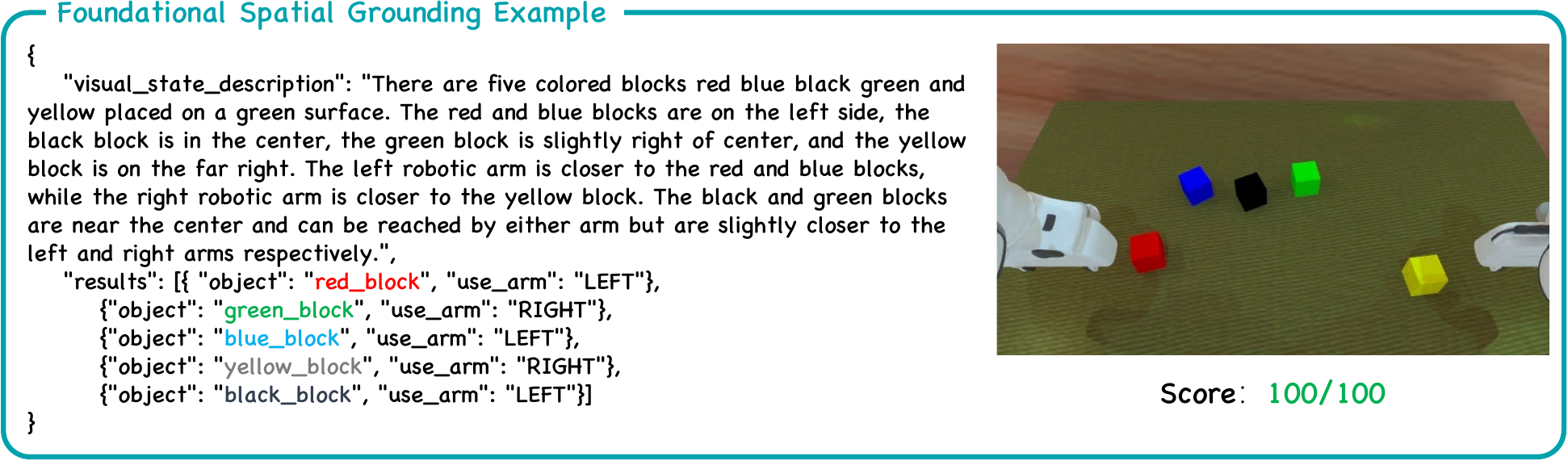}
    \caption{High-quality reasoning example.}
    \label{fig:example_spatial1}
\end{figure}
\begin{figure}[H]
    \centering
    \includegraphics[width=0.8\textwidth]{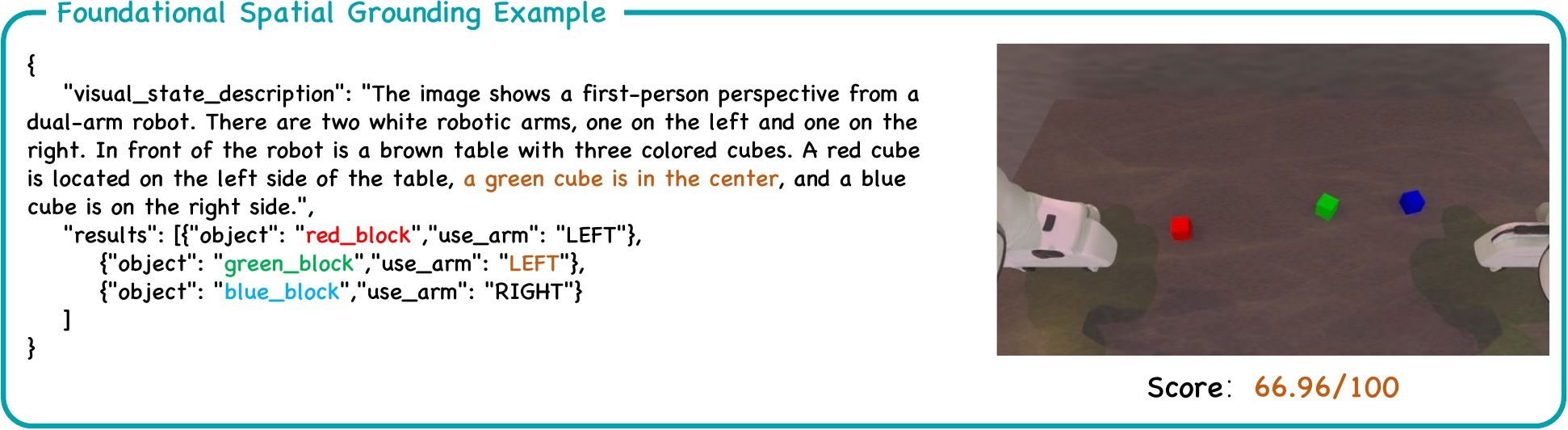}
    \caption{Average-quality reasoning example with spatial ambiguity.}
    \label{fig:example_spatial2}
\end{figure}
\begin{figure}[H]
    \centering
    \includegraphics[width=0.8\textwidth]{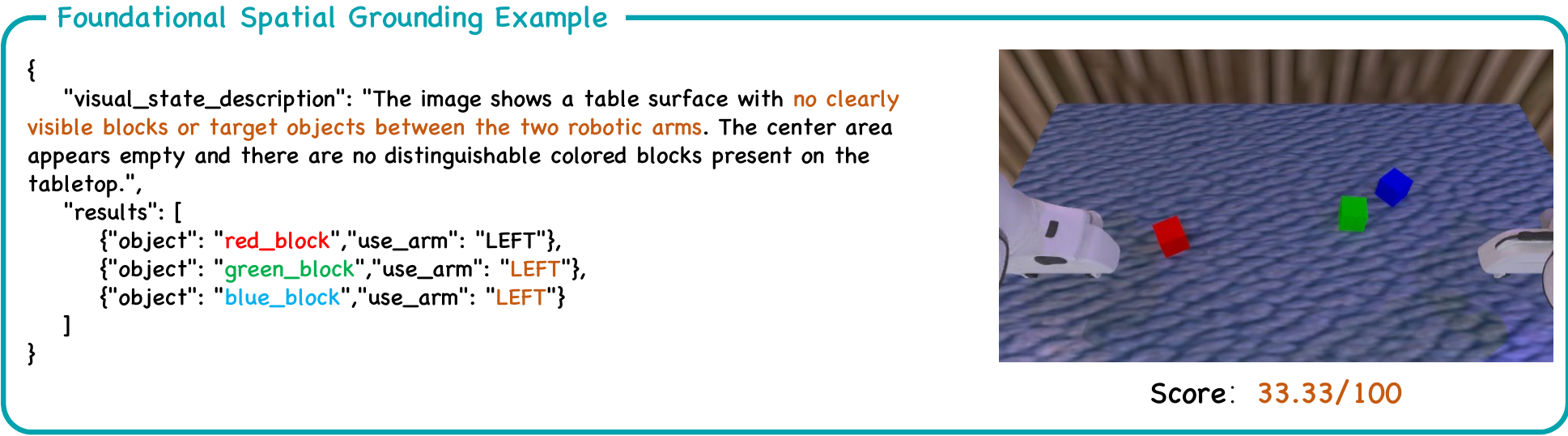}
    \caption{Low-quality reasoning example with significant visual hallucination.}
    \label{fig:example_spatial3}
\end{figure}

\subsection{Strategic Coordination Planning}
\label{app:high_level_task_details}

\medskip\noindent\textbf{Task Introduction}

The task ID, full name, and description (instructions provided to the planner) are shown in Table~\ref{tab:highlevel_task_list_table}.
\begin{table}[htbp]
    \centering
    \caption{List of 14 Strategic Coordination Planning Manipulation Tasks}
    \label{tab:highlevel_task_list_table}
    \begin{tabular}{l l p{10cm}} 
        \toprule
        \textbf{ID} & \textbf{Full Task Name} & \textbf{Description} \\
        \midrule
        \multicolumn{3}{c}{\textit{Independent Parallel Manipulation Tasks}} \\
        \midrule
        Place1 & \texttt{place\_cans\_plasticbox} &Use dual arm to pick and place cans into plasticbox.  \\ \addlinespace[2ex]
        Place2 & \texttt{blocks\_cross\_shape} & There are five blocks on the table, the color of the blocks is random, move the blocks to the center of the table, and arrange them to a cross shape. The red block should be on the left. The black block should be on the center. The blue block should be on the right. The green should be inside. The yellow should be closest to you. \\ \addlinespace[2ex]
        Ranking1 & \texttt{blocks\_ranking\_size} & There are three blocks on the table, the color of the blocks is random, move the blocks to the center of the table, and arrange them from largest to smallest, from left to right. \\\addlinespace[2ex]
        Ranking2 & \texttt{blocks\_ranking\_rgb} & Place the red block, green block, and blue block in the order of red, green, and blue from left to right, placing in a row.  \\\addlinespace[2ex]
        Stack1 & \texttt{stack\_blocks\_three} & There are three blocks on the table, the color of the blocks is red, green and blue, move the blocks to the center of the table, and stack the blue block on the green block, and the green block on the red block. \\\addlinespace[2ex]
        Stack2 & \texttt{stack\_bowls\_three} & Stack the three bowls on top of each other.  \\
        \midrule
        \multicolumn{3}{c}{\textit{Sequential Collaborative Manipulation Tasks}} \\
        \midrule
        Handover1 & \texttt{handover\_mic} & Use one arm to grasp the microphone on the table and handover it to the other arm. \\ \addlinespace[2ex]
        Handover2 & \texttt{handover\_block} & Use the left arm to grasp the red block on the table, handover it to the right arm and place it on the blue pad. \\\addlinespace[2ex]
        Handover3 & \texttt{hanging\_mug} & Use left arm to pick the mug on the table, rotate the mug and put the mug down in the middle of the table, use the right arm to pick the mug and hang it onto the rack.  \\\addlinespace[2ex]
        Place3 & \texttt{place\_burger\_fries} & Use dual arm to pick the hamburg and frenchfries and put them onto the tray. \\\addlinespace[2ex]
        Place4 & \texttt{place\_object\_basket} & Use one arm to grab the target object and put it in the basket, then use the other arm to grab the basket, and finally move the basket slightly away. \\\addlinespace[2ex]
        Place5 & \texttt{place\_bread\_skillet} & If there is one bread on the table, use one arm to grab the bread and put it into the skillet. \\\addlinespace[2ex]
        Place6 & \texttt{blocks\_tower} & There are four blocks on the table, the color of the blocks is random, move the blocks to the center of the table, and stack them to a block tower from largest to smallest, from bottom to top. \\\addlinespace[2ex]
        Place7 & \texttt{put\_bottles\_dustbin} & Use arms to grab the bottles and put them into the dustbin to the left of the table. \\
        \bottomrule
    \end{tabular}
\end{table}

\medskip\noindent\textbf{Manipulator Allocation Feedback and Truncation Mechanism}

In the Strategic Coordination Planning tier, successful task completion relies on both logical sequencing and correct arm assignment based on workspace reachability. To ensure a rigorous evaluation of the model's logical reasoning and temporal planning capabilities without them being disproportionately bottlenecked by fundamental spatial reasoning failures, we implement an automated Manipulator Allocation Feedback and Truncation mechanism at the interface layer.

During the execution pipeline, the interface layer intercepts the model’s structured JSON output, specifically the arm\_tag parameter (``left'' or ``right''). The system verifies the validity of this assignment against the simulator's ground-truth kinematic constraints and object reachability. If the model selects an execution arm that is kinematically unable to reach the target or likely to cause a singularity, the system truncates the action before physical execution. Instead of terminating the episode as a failure, the environment provides a closed-loop textual feedback to the model (\textit{e.g.}, ``Action failed: Action Failed: target green\_block is too far, right arm can not finish this `grasp' action! Please use another arm!''). This allows the agent to re-observe the state and replan its execution strategy within the same episode.

The primary motivation for this mechanism is to decouple Foundational Spatial Grounding from high-level Strategic Coordination Planning. By providing an automated ``guardrail'' for arm allocation, we can more accurately assess whether an MLLM understands the logical dependencies of complex tasks—such as the sequential requirements in Handover\_block or Place\_bread\_skillet—rather than penalizing its planning performance solely due to a one-time spatial miscalculation.

It is important to note that this truncation mechanism is not exhaustive. It primarily targets clear kinematic violations and reachability issues. More nuanced errors—such as ``model illusions'' during complex bimanual coordination or subtle workspace overlaps—may still bypass the filter. Consequently, a small percentage of End-Effector Allocation Errors remains present in our final Error Analysis, providing a realistic reflection of the coupling between spatial awareness and high-level planning in dual-arm scenarios.

\newpage
\medskip\noindent\textbf{Skills Set}

\label{appendix:skills}

\begin{skillbox}
\begin{description}[leftmargin=0pt, itemsep=1em]

\item[Grasp Skill (\texttt{grasp\_actor})] 
Picks up a specified object using the selected arm.
\begin{itemize}[label=$\diamond$]
    \item \textbf{Parameters:} 
    \textit{actor}: target object to grasp; 
    \textit{arm\_tag}: arm selection; 
    \textit{pre\_grasp\_dis}: distance to the pre-grasp waypoint (default 0.1\,m); 
    \textit{grasp\_dis}: approach distance from the pre-grasp pose (default 0\,m); 
    \textit{gripper\_pos}: gripper closing position (default fully closed); 
    \textit{contact\_point\_id} (optional): indices of candidate contact points.
    \item \textbf{Description:} The arm first moves to a pre-grasp position offset from the object, then approaches the grasp pose and closes the gripper. If no contact point is specified, the grasp point is selected automatically.
\end{itemize}

\item[Place Skill (\texttt{place\_actor})] 
Places a currently held object at a specified target pose.
\begin{itemize}[label=$\diamond$]
    \item \textbf{Parameters:} 
    \textit{actor}: grasped object; 
    \textit{arm\_tag}: arm holding the object; 
    \textit{target\_pose}: target position and orientation (recommended as xyz + quaternion); 
    \textit{pre\_dis}: pre-placement offset distance (default 0.1\,m); 
    \textit{dis}: final placement distance (default 0.02\,m); 
    \textit{is\_open}: whether to open the gripper after placement.
    \item \textbf{Optional arguments:} 
    alignment strategy (\textit{free}, \textit{align}, \textit{auto}); 
    alignment axes; 
    pre-placement direction.
    \item \textbf{Description:} The arm moves the object to a pre-placement pose, optionally aligns it with the target orientation, then performs a guarded approach to the final placement pose and releases the object if specified.
\end{itemize}

\item[Relative Motion (\texttt{move\_by\_displacement})] 
Moves the end-effector along relative directions.
\begin{itemize}[label=$\diamond$]
    \item \textbf{Parameters:} 
    \textit{arm\_tag}: arm selection; 
    \textit{x}, \textit{y}, \textit{z}: Cartesian displacement in meters; 
    \textit{quat} (optional): target orientation; 
    \textit{move\_axis}: reference frame (\textit{world} or \textit{arm}).
    \item \textbf{Description:} Translates the end-effector by the specified displacement under the chosen reference frame, with optional orientation control.
\end{itemize}

\item[Absolute Motion (\texttt{move\_to\_pose})] 
Moves the end-effector to an absolute target pose.
\begin{itemize}[label=$\diamond$]
    \item \textbf{Parameters:} 
    \textit{arm\_tag}: arm selection; 
    \textit{target\_pose}: absolute position and orientation.
    \item \textbf{Description:} Plans and executes a collision-aware trajectory to reach the target pose.
\end{itemize}

\item[Gripper Control (\texttt{open\_gripper} / \texttt{close\_gripper})] 
Opens or closes the gripper of the specified arm.
\begin{itemize}[label=$\diamond$]
    \item \textbf{Parameters:} 
    \textit{arm\_tag}: arm selection; 
    \textit{pos}: gripper position (0 for fully closed, 1 for fully open).
\end{itemize}

\item[Home Reset (\texttt{back\_to\_origin})] 
Returns the arm to its predefined initial configuration.
\begin{itemize}[label=$\diamond$]
    \item \textbf{Parameters:} 
    \textit{arm\_tag}: arm to reset.
    \item \textbf{Description:} Moves the arm back to a safe home pose to clear the workspace.
\end{itemize}

\end{description}
\end{skillbox}

\newpage
\medskip\noindent\textbf{Examples}

\begin{figure}[H]
    \centering
    \includegraphics[width=0.8\textwidth]{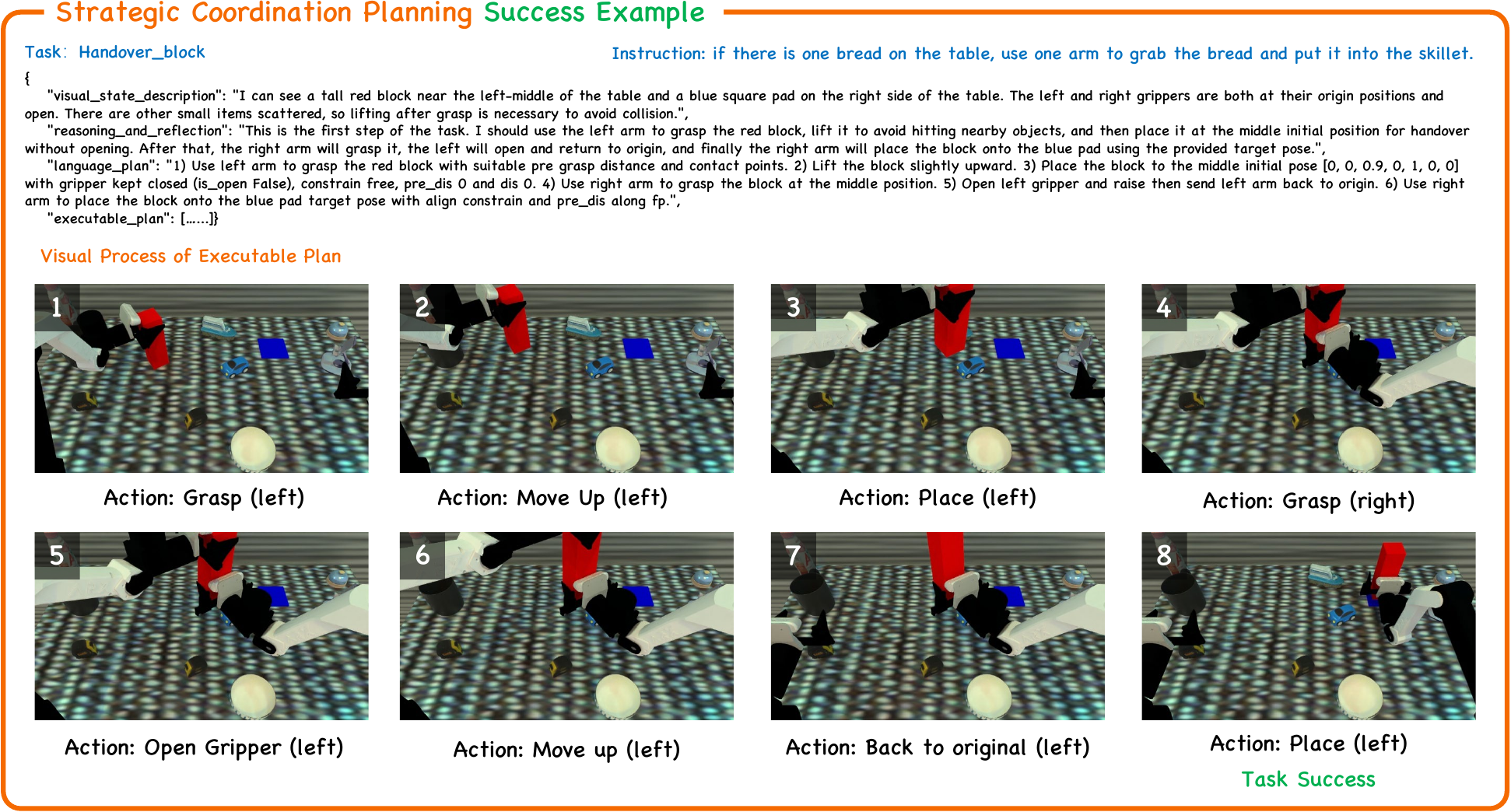}
    \caption{Strategic Coordination Planning Success example of Gemini-2.5-Pro: Handover\_block.}
    \label{fig:example_highlevel1}
\end{figure}

\input{appendix/examples/highlevel_example}

\begin{figure}[H]
    \centering
    \includegraphics[width=0.8\textwidth]{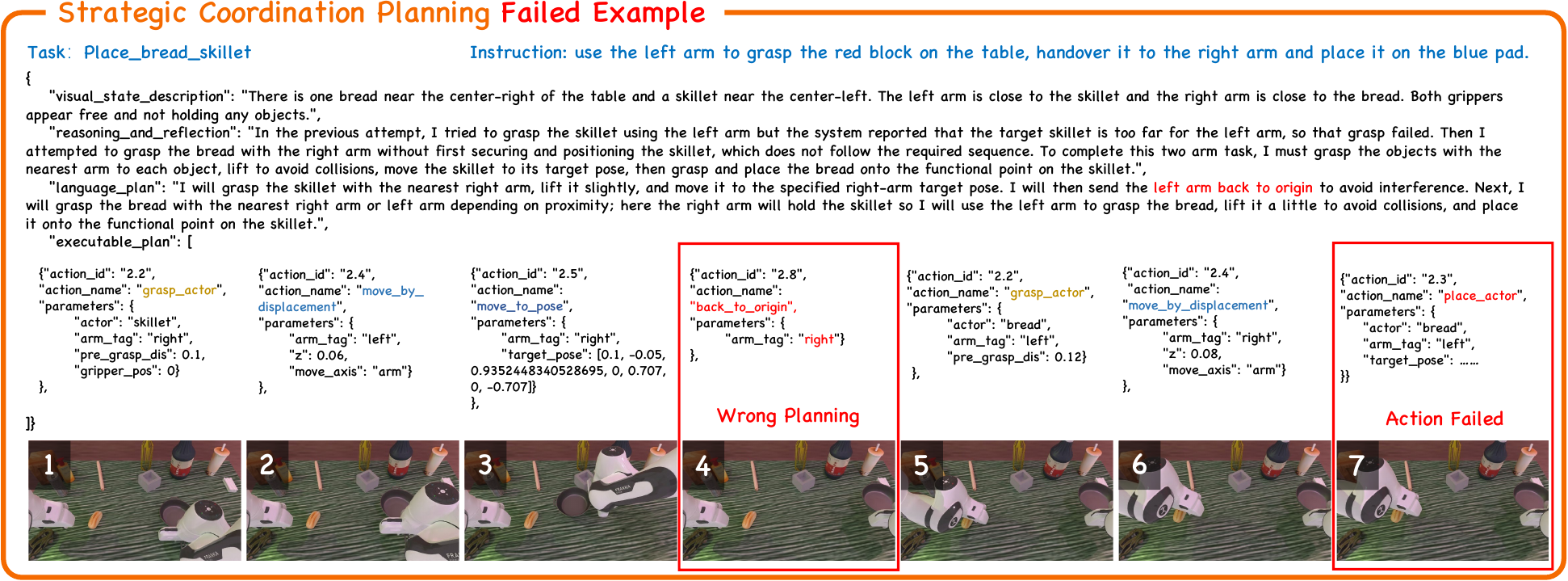}
    \caption{Strategic Coordination Planning Failed example of GPT-5: Place\_bread\_skillet.}
    \label{fig:example_highlevel2}
\end{figure}

\clearpage
\subsection{Fine-Grained Action Control}
\label{app:low_level_task_details}

\medskip\noindent\textbf{Task Introduction}

The task ID, full name, and description (instructions provided to the planner) are shown in Table~\ref{tab:lowlevel_task_list_table}.
\begin{table}[htbp]
    \centering
    \caption{List of 5  Fine-Grained Action Control Manipulation Tasks}
    \label{tab:lowlevel_task_list_table}
    \begin{tabular}{l l p{2cm} p{7.5cm}} 
        \toprule
        \textbf{ID} & \textbf{Full Task Name} & \textbf{Class} & \textbf{Description} \\
        \midrule     
        Place8 & \texttt{place\_object\_scale}& Independent Manipulation (single arm) & Use one arm to grab the object and put it on the scale. \\\addlinespace[2ex]
        Place9 & \texttt{place\_burger\_fries}&Sequential Collaborative Manipulation & Use dual arm to pick the hamburg and frenchfries and put them onto the tray. \\\addlinespace[2ex]
        Place10 & \texttt{place\_bread\_skillet}&Sequential Collaborative Manipulation & If there is one bread on the table, move skillet to center and use one arm to grab the bread and put it into the skillet. \\\addlinespace[2ex]
        Grab1 & \texttt{grab\_roller}& Synchronous Collaborative Manipulation & Use both arms to grab the roller on the table and lift it to $z>0.8$(m). \\\addlinespace[2ex]
        Stack3 & \texttt{stack\_blocks\_two}& Independent Parallel Manipulation & There are two blocks on the table, the color of the blocks is red, green, move the blocks to the center of the table, and stack the blocks. \\
        \bottomrule
    \end{tabular}
\end{table}

\newpage
\medskip\noindent\textbf{Examples}

\begin{figure}[H]
    \centering
    \includegraphics[width=0.8\textwidth]{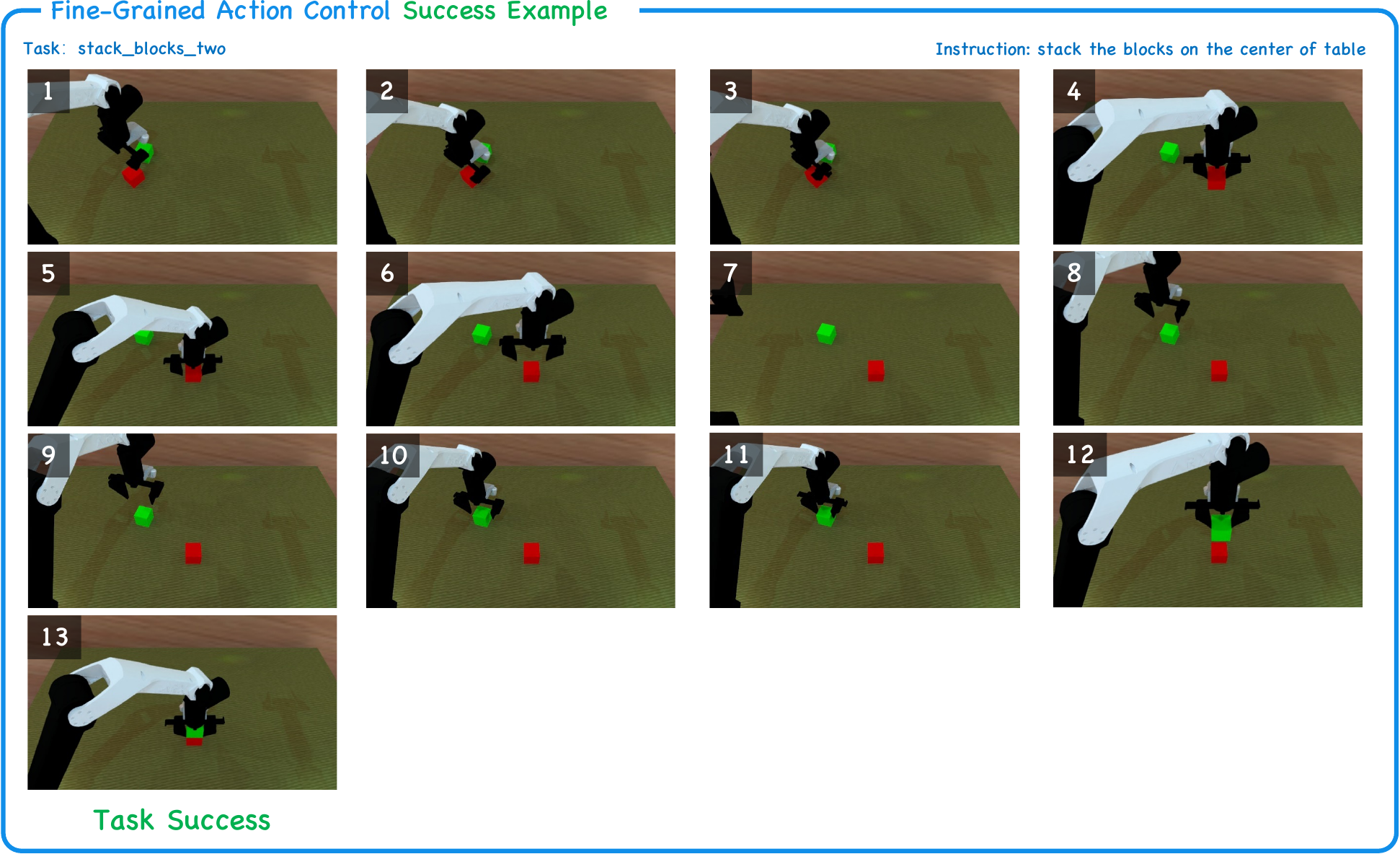}
    \caption{Fine-Grained Action Control Success example of GPT-5: Stack\_blocks\_two.}
    \label{fig:example_lowlevel1}
\end{figure}

\newpage
\input{appendix/examples/lowlevel_example}

\begin{figure}[H]
    \centering
    \includegraphics[width=0.8\textwidth]{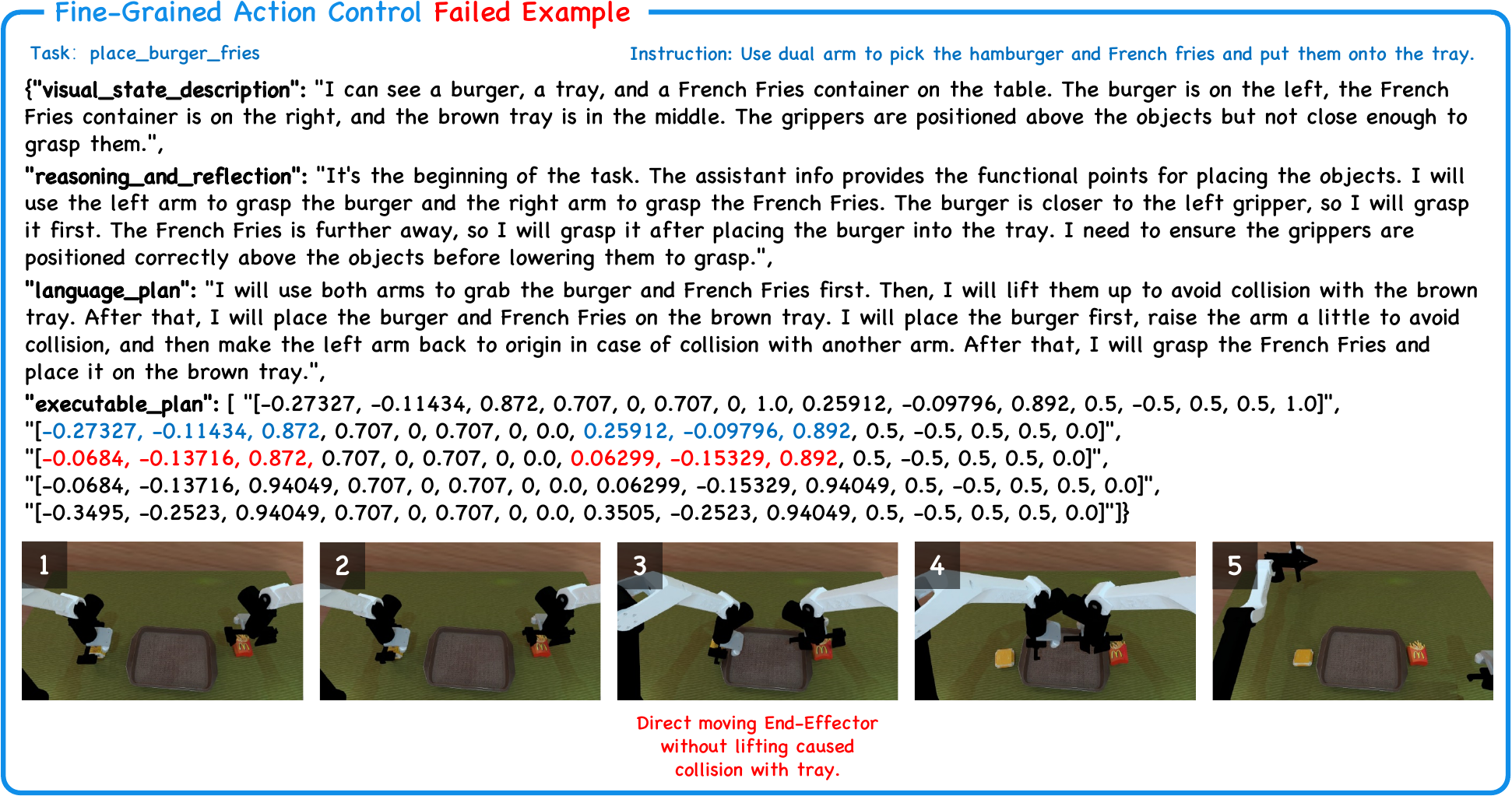}
    \caption{Fine-Grained Action Control Failed example of InternVL3-78B: Place\_burger\_fries.}
    \label{fig:example_lowlevel2}
\end{figure}
\section{Error Analysis}
\label{app:error_analysis}
We first divide the errors we encounter into two main categories: environmental errors and model output errors.
\subsection{Environmental Errors}
One type of environmental error occurs when the robotic arm's actual behavior fails to meet the documentation's expectations when the model calls relevant APIs through the simulation environment interface. This problem is not caused by the decision-making layer model, but rather by a flaw in the internal function settings of the simulation environment at the execution layer. Since different tasks involve different calls to the simulation environment interface, the probability of this type of error varies across tasks. However, within the same task, the occurrence of this type of error is independent of the specific model used by the decision-making layer, occurring with approximately the same probability for all models.

An example is grasp problem. In the simulation environment, all objects are generated with one face always pointing vertically upwards. If an object is hit by a collision, this face may turn in another direction. For a geometrically symmetrical cube, this should not affect its grasping ability. However, because the simulation environment implicitly depends on the object's orientation when calling the grasping function, the grasping action cannot be completed when the object's orientation changes.

Another type is the scene randomization problem in certain episodes. When randomizing the scene using a random seed, some interfering objects are generated in inappropriate locations, severely interfering with normal operation. Since all models use the same random seed and scene for each episode during testing, all models will encounter the scene randomization problem and will be unable to complete the task in some task episodes.
\subsection{Model Output Errors}
We categorize model output errors into three types: Perceptual Reasoning Error, Strategic Planning Error, and Format Error. Since Format Error only occurs in models with a small number of parameters(usually less than 10B), our analysis focuses on the following two core categories:

\medskip\noindent\textbf{1) Perceptual Reasoning Error.}
This category focuses on inference errors following visual observation. It reflects the model's failure to correctly understand the "current state." It is further divided into:
\begin{itemize}
    \item \textbf{State Estimation Misjudgment.}  This error primarily occurs when the model makes inferences based on current observations, encountering problems in inferring the state and progress of the current task. An example is when the model, after having already placed a block in the target location, continues to repeatedly grab and place that block without addressing other unfinished tasks.
    \item \textbf{End-Effector Allocation Error.} This error primarily arises when the model, based on current observations, encounters problems in determining whether to use the left or right actuator in the current state. This issue stems from insufficient spatial capabilities in a dual-arm environment, or significant model illusion during the end-effector allocation process in dual-arm scenarios. This capability was tested within the first part of our Foundational Spatial Grounding section. Therefore, to better evaluate the model's capabilities in other areas, we automatically truncated most incorrect End-Effector Allocation cases in the interface layer between the model output and the simulation environment, providing feedback to the model that the current end-effector cannot complete the action until the model adopts the correct end-effector. However, this automatic truncation cannot capture all incorrect End-Effector Allocation scenarios; therefore, some End-Effector Allocation Errors will still occur, affecting dual-arm operation tasks under Strategic Coordination Planning.
    \item \textbf{Physical Attribute Misreasoning.} This error mainly occurs when the model makes inferences based on current observations, and it has a problem in inferring the size of objects. It mainly appears in two tasks that require determining the size of blocks.
\end{itemize} 
\medskip\noindent\textbf{2) Strategic Planning Error.}
This category involves errors in the subsequent action sequence planning, even if the environment is correctly perceived. It includes:
\begin{itemize}
    \item \textbf{Action Sequencing Error.} This error primarily stems from the model's failure to output the correct action sequence when planning actions, based on the specific task requirements or the task flow hinted at in the assistant info. A typical example is in dual-arm manipulation tasks: after grasping an object, the end effector must be raised before placing it, and after placement, the end effector must be raised again before the robotic arm returns to its original position; otherwise, the end effector's movement may collide with other objects on the table. This error also frequently occurs in object-transfer tasks, where the order of actions is crucial. If the model fails to output the correct action sequence, the transfer will fail.
    \item \textbf{Bimanual Conflict/Collision.} This error primarily stems from the model's failure to consider the conflict between the two robotic arms during action planning. This leads to collisions or disruptions in action execution due to these conflicts. This error is often accompanied by a State Estimation Misjudgment Error because the model's previous output may not have properly positioned the robotic arms. If problems arise in inferring the current task's state and progress, and the robotic arms are not properly positioned beforehand, subsequent action execution is prone to Bimanual Conflict/Collision.
    \item \textbf{Action Parameter Inconsistency.} This error occurs because the model failed to output task-specific action parameters when formatting the action output in JSON format. In a task, the parameters that should be set for different scenarios and types of actions are usually indicated in the prompt; however, some models still occasionally produce incorrect action parameter outputs.
\end{itemize}

\section{Visual Observation Setup}
\label{app:visual_setup}

\begin{figure}[h]
    \centering
    \begin{subfigure}{0.48\linewidth}
        \centering
        \includegraphics[width=\linewidth]{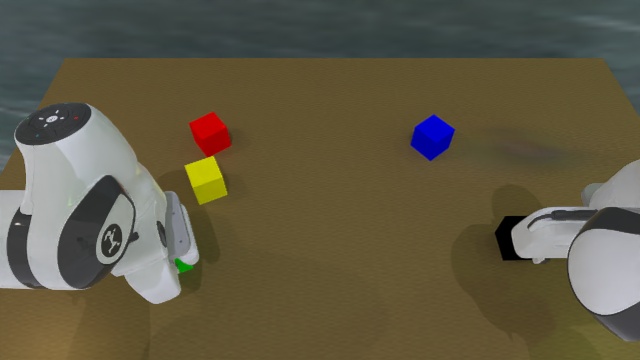} 
        \caption{Ego-centric View (Primary)}
        \label{fig:view_ego}
    \end{subfigure}
    \hfill
    \begin{subfigure}{0.48\linewidth}
        \centering
        \includegraphics[width=\linewidth]{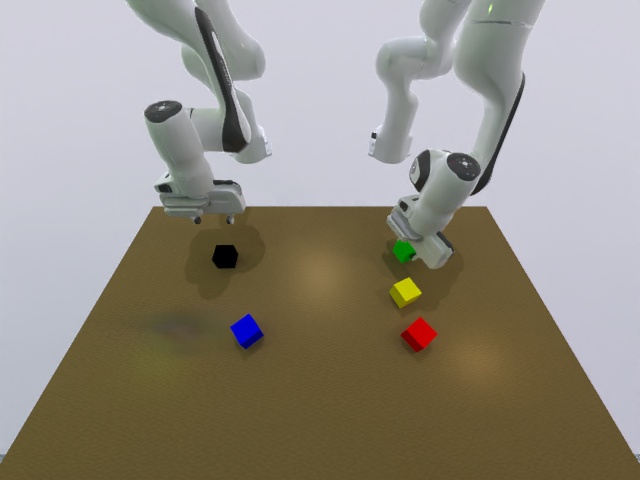} %
        \caption{Third-person View (Auxiliary)}
        \label{fig:view_third}
    \end{subfigure}
    
    \caption{\textbf{Visualization of Observation Space.} 
    The left panel (a) shows the ego-centric view subject to occlusion during interaction. 
    The right panel (b) displays the third-person view, which provides global context and is utilized to mitigate occlusion in Strategic Coordination Planning and selected Fine-Grained Action Control tasks.}
    \label{fig:view_comparison}
\end{figure}

In our experiments, the agent receives visual inputs from different perspectives depending on the task configuration. 
In Strategic Coordination Planning, we provide images from both perspectives to the MLLM planner. 
In Fine-Grained Action Control, we primarily utilize only the \textbf{ego-centric view}, while the \textbf{third-person view} is investigated in our ablation study.
Fig~\ref{fig:view_comparison} illustrates the comparison between these two observation modes.

\begin{itemize}
    \item \textbf{Ego-centric View (Primary):} Mounted on the robot's end-effector (or head), providing detailed visual feedback for precise manipulation but suffering from occlusion when the gripper is close to the object.
    
    \item \textbf{Third-person View:} A fixed external camera that provides a global overview of the workspace. This view is crucial for resolving spatial ambiguities (\textit{e.g.}, occlusion) and is used as a default input for Strategic Coordination Planning. However, it introduces two challenges for visual reasoning:
    \begin{enumerate}
        \item \textbf{Reduced Object Resolution:} Due to the greater camera distance, manipulable objects occupy fewer pixels compared to the ego-centric view.
        \item \textbf{Spatial Inversion:} The perspective creates a counter-intuitive visual reversal where the robot's physical \textit{right arm} appears on the \textit{left side} of the image, and the \textit{left arm} on the \textit{right}.
    \end{enumerate}
    For models with limited capabilities, these factors can act as visual interference rather than helpful context, potentially leading to the performance degradation observed in our Ablation Study.
\end{itemize}
\section{Prompt Template}
\label{app:prompt}
\subsection{Foundational Spatial Grounding Prompt Template}
This is the prompt template for sparse and cluttered settings in Foundational Spatial Grounding.
The dense setting is nearly the same, except for the final output example, as it involves five blocks.

\input{appendix/prompts/spatial_prompt}

\subsection{Strategic Coordination Planning Prompt Template}
This is a Strategic Coordination Planning prompt template for the blocks\_ranking\_rgb task. The following prompt is used at the beginning of each task execution.

\input{appendix/prompts/highlevel_prompt}

We also provide output examples in image form to assist the model with action planning.

\begin{figure}[H]
    \centering
    \includegraphics[width=0.8\textwidth]{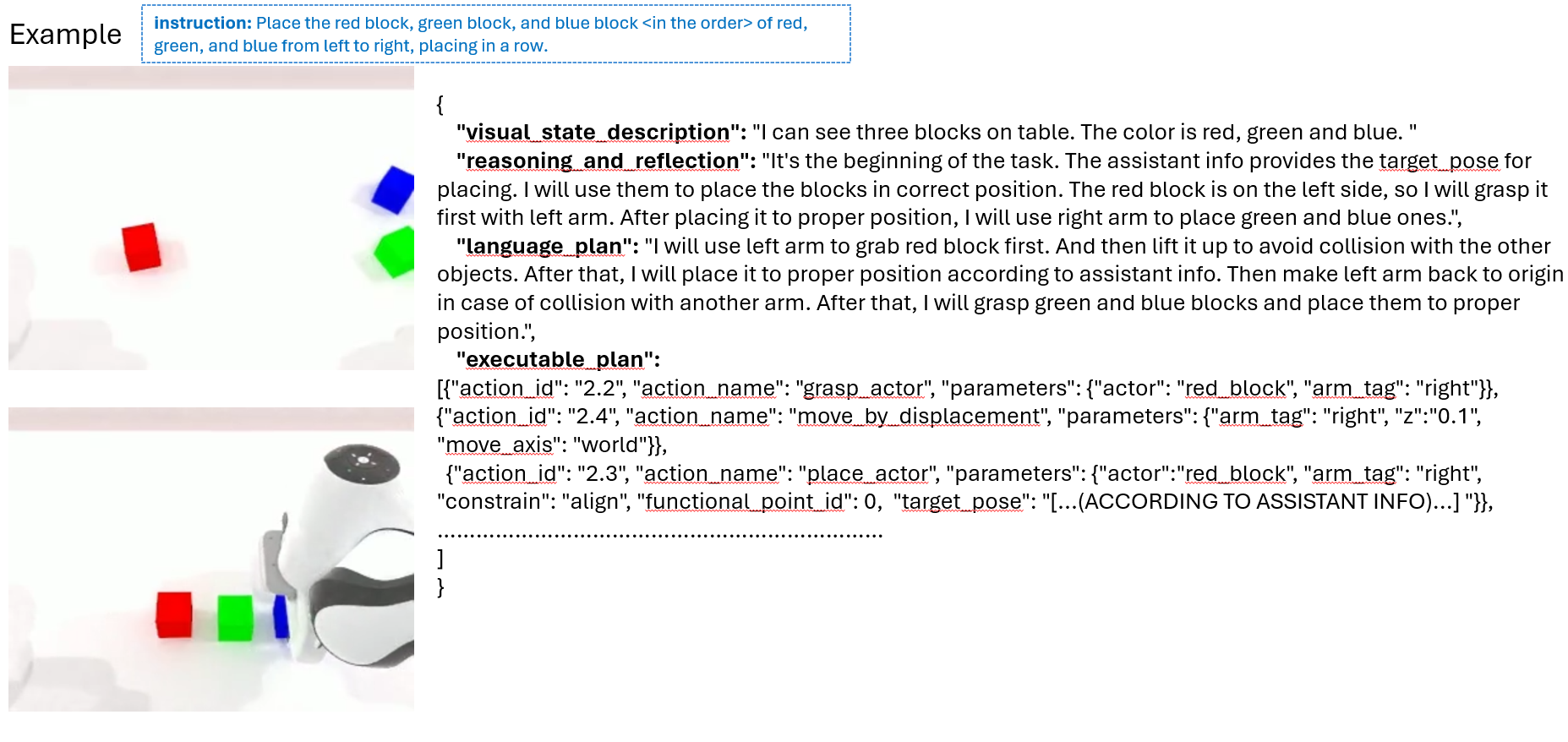}
    \caption{Strategic Coordination Planning example in img form: Blocks\_ranking\_rgb.}
    \label{fig:example_highlevel4}
\end{figure}

\subsection{Fine-Grained Action Control Prompt Template}
\input{appendix/prompts/lowlevel_prompt}

%% file: appendix/examples/highlevel_example.tex
\begin{lstlisting}[language=json, caption={Planner Output of High-Level Action Planning Success example: Handover\_block.}, label={lst:planner-json},
  numbers=none,escapeinside={(*@}{@*)},captionpos=b]
(*@\textbf{Output Step 1:}@*)
{
    (*@\textbf{"visual\_state\_description":}@*) "I can see a tall red block near the left-middle of the table and a blue square pad on the right side of the table. The left and right grippers are both at their origin positions and open. There are other small items scattered, so lifting after grasp is necessary to avoid collision.",
    (*@\textbf{"reasoning\_and\_reflection":}@*) "This is the first step of the task. I should use the left arm to grasp the red block, lift it to avoid hitting nearby objects, and then place it at the middle initial position for handover without opening. After that, the right arm will grasp it, the left will open and return to origin, and finally the right arm will place the block onto the blue pad using the provided target pose.",
    (*@\textbf{"language\_plan":}@*) "1) Use left arm to grasp the red block with suitable pre grasp distance and contact points. 2) Lift the block slightly upward. 3) Place the block to the middle initial pose [0, 0, 0.9, 0, 1, 0, 0] with gripper kept closed (is_open False), constrain free, pre_dis 0 and dis 0. 4) Use right arm to grasp the block at the middle position. 5) Open left gripper and raise then send left arm back to origin. 6) Use right arm to place the block onto the blue pad target pose with align constrain and pre_dis along fp.",
    (*@\textbf{"executable\_plan":}@*) [
        {
            "action_id": "2.2",
            "action_name": "grasp_actor",
            "parameters": {
                "actor": "block",
                "arm_tag": "left",
                "pre_grasp_dis": 0.07,
                "grasp_dis": 0,
                "gripper_pos": 0,
                "contact_point_id": [0, 1, 2, 3]
            }
        },
        {
            "action_id": "2.4",
            "action_name": "move_by_displacement",
            "parameters": {
                "arm_tag": "left",
                "x": 0,
                "y": 0,
                "z": 0.08,
                "move_axis": "world"
            }
        },
        {
            "action_id": "2.3",
            "action_name": "place_actor",
            "parameters": {
                "actor": "block",
                "arm_tag": "left",
                "target_pose": [0, 0, 0.9, 0, 1, 0, 0],
                "functional_point_id": 0,
                "pre_dis": 0,
                "dis": 0,
                "is_open": false,
                "kwargs": {
                    "constrain": "free",
                    "pre_dis_axis": "fp"
                }
            }
        },
        {
            "action_id": "2.2",
            "action_name": "grasp_actor",
            "parameters": {
                "actor": "block",
                "arm_tag": "right",
                "pre_grasp_dis": 0.07,
                "grasp_dis": 0,
                "gripper_pos": 0,
                "contact_point_id": [4, 5, 6, 7]
            }
        },
        {
            "action_id": "2.7",
            "action_name": "open_gripper",
            "parameters": {
                "arm_tag": "left",
                "pos": 1
            }
        },
        {
            "action_id": "2.4",
            "action_name": "move_by_displacement",
            "parameters": {
                "arm_tag": "left",
                "x": 0,
                "y": 0,
                "z": 0.06,
                "move_axis": "world"
            }
        },
        {
            "action_id": "2.8",
            "action_name": "back_to_origin",
            "parameters": {
                "arm_tag": "left"
            }
        },
        {
            "action_id": "2.3",
            "action_name": "place_actor",
            "parameters": {
                "actor": "block",
                "arm_tag": "right",
                "target_pose": [0.238135, 0.160577, 0.730889, 0, 1, 0, 0],
                "functional_point_id": 0,
                "pre_dis": 0.05,
                "dis": 0,
                "is_open": true,
                "kwargs": {
                    "constrain": "align",
                    "pre_dis_axis": "fp"
                }
            }
        }
    ]
}
\end{lstlisting}

%% file: appendix/examples/lowlevel_example.tex
\begin{lstlisting}[language=json, caption={Planner output for a dual-arm stacking task}, label={lst:planner-json}, 
  numbers=none,escapeinside={(*@}{@*)},captionpos=b]
(*@\textbf{Output Step 1:}@*)

{
    (*@\textbf{"visual\_state\_description":}@*) "There are two cubes on the table: a red cube (block1) at approximately x=-0.199, y=-0.071 and a green cube (block2) at approximately x=-0.174, y=0.029. Both are on the left side of the table relative to the robot. The left and right end effectors are at their origin poses above the table edges with grippers open.",
    (*@\textbf{"reasoning\_and\_reflection":}@*) "Both cubes are on the left side, so the left arm should pick the nearer one first. The red cube (block1) at y=-0.071 is slightly closer to the target center y=-0.13, so I will grasp block1 with the left arm, place it at center as the bottom block, then move the left arm up and back to origin to avoid collision. Next, I will use the left arm again for block2 because it is still on the left side, while keeping the right arm parked high at origin so it does not interfere. Grasp orientations for blocks must use the provided quaternions. When placing at center, use orientation [0.5,-0.5,0.5,0.5]. Heights: approach above at safe z about 0.95 to 1.0, then lower to grasp around object height plus gripper offset (0.73949 + 0.162 (*@$\approx$@*) 0.90149, use 0.9). For placing bottom block, target z should be 0.7244852714832615 + 0.162 = 0.88649; use 0.8865. For stacking top block, target z should be 0.7744852714832615 + 0.162 = 0.93649; use 0.9365. I will close gripper to pick, move to center, release, then raise and repeat for the second block. Keep right arm high and at origin throughout.",
    (*@\textbf{"language\_plan":}@*) "1) Left arm move above block1, descend, close to grasp. 2) Lift and place at center as bottom, release. 3) Raise left arm to 1.08 and move back to origin. 4) Move left arm again above block2, grasp. 5) Move to center above placed block and place on top, release. 6) Raise and return to origin. Right arm stays high and open to avoid collision.",
    (*@\textbf{"executable\_plan":}@*) [
        "[-0.20, -0.07, 0.98, 0.64743, -0.2843, 0.64743, 0.2843, 1.0, 0.3505, -0.2523, 1.08, 0.70711, -0.00001, 0.00001, 0.70711, 1.0]",
        "[-0.20, -0.07, 0.90, 0.64743, -0.2843, 0.64743, 0.2843, 1.0, 0.3505, -0.2523, 1.08, 0.70711, -0.00001, 0.00001, 0.70711, 1.0]",
        "[-0.20, -0.07, 0.90, 0.64743, -0.2843, 0.64743, 0.2843, 0.0, 0.3505, -0.2523, 1.08, 0.70711, -0.00001, 0.00001, 0.70711, 1.0]",
        "[-0.01, -0.13, 0.98, 0.5, -0.5, 0.5, 0.5, 0.0, 0.3505, -0.2523, 1.08, 0.70711, -0.00001, 0.00001, 0.70711, 1.0]",
        "[0.0, -0.13, 0.8865, 0.5, -0.5, 0.5, 0.5, 1.0, 0.3505, -0.2523, 1.08, 0.70711, -0.00001, 0.00001, 0.70711, 1.0]",
        "[0.0, -0.13, 0.98, 0.5, -0.5, 0.5, 0.5, 1.0, 0.3505, -0.2523, 1.08, 0.70711, -0.00001, 0.00001, 0.70711, 1.0]",
        "[-0.3495, -0.2523, 1.08, 0.70711, -0.00001, 0.00001, 0.70711, 1.0, 0.3505, -0.2523, 1.08, 0.70711, -0.00001, 0.00001, 0.70711, 1.0]",
        "[-0.174, 0.029, 0.98, 0.56669, -0.42293, 0.56669, 0.42293, 1.0, 0.3505, -0.2523, 1.08, 0.70711, -0.00001, 0.00001, 0.70711, 1.0]",
        "[-0.174, 0.029, 0.90, 0.56669, -0.42293, 0.56669, 0.42293, 1.0, 0.3505, -0.2523, 1.08, 0.70711, -0.00001, 0.00001, 0.70711, 1.0]",
        "[-0.174, 0.029, 0.90, 0.56669, -0.42293, 0.56669, 0.42293, 0.0, 0.3505, -0.2523, 1.08, 0.70711, -0.00001, 0.00001, 0.70711, 1.0]",
        "[0.0, -0.13, 0.98, 0.5, -0.5, 0.5, 0.5, 0.0, 0.3505, -0.2523, 1.08, 0.70711, -0.00001, 0.00001, 0.70711, 1.0]",
        "[0.0, -0.13, 0.9365, 0.5, -0.5, 0.5, 0.5, 1.0, 0.3505, -0.2523, 1.08, 0.70711, -0.00001, 0.00001, 0.70711, 1.0]",
        "[0.0, -0.13, 1.02, 0.5, -0.5, 0.5, 0.5, 1.0, 0.3505, -0.2523, 1.08, 0.70711, -0.00001, 0.00001, 0.70711, 1.0]",
        "[-0.3495, -0.2523, 1.08, 0.70711, -0.00001, 0.00001, 0.70711, 1.0, 0.3505, -0.2523, 0.94049, 0.70711, -0.00001, 0.00001, 0.70711, 1.0]"
    ]
}
(*@\textbf{Output Step 2:}@*)

{
    (*@\textbf{"visual\_state\_description":}@*) "A dual-arm robot is above a table with two blocks: a green block near the left arm at approximately [-0.174, 0.029, 0.739] and a red block near the center-right at approximately [0.002, -0.124, 0.739]. The left gripper appears close to the green block and the gripper is open. The right arm is parked high around [0.3505, -0.2523, 1.08] and is open.",
    (*@\textbf{"reasoning\_and\_reflection":}@*) "The goal is to move both blocks to the table center at [0, -0.13] and stack them. The action history shows multiple approach and open/close attempts but no successful grasp or placement; both actions were successful in execution but did not complete the task because the grippers likely did not close on the blocks at the correct height and orientation, and there was no sequence that moved and released at the center. The current positions indicate the green block is under the left arm and the red block is slightly in front of center. To avoid collisions, I will use the left arm to pick the green block (block2) with quat [0.56668, -0.42293, 0.56668, 0.42293]. I will use the right arm to pick the red block (block1) with quat [0.49911, -0.50089, 0.49911, 0.50089]. When placing at the center, I will use orientation [0.5, -0.5, 0.5, 0.5]. I must account for the 0.162 m gripper height offset and place the bottom block gripper z at 0.7244852715 + 0.162 = 0.88649, and the top block gripper z at 0.7744852715 + 0.162 = 0.93649. I will first stack the green block as the bottom using the left arm, then stack the red block on top using the right arm. I will keep the non-operating arm raised and away to avoid collision.",
    (*@\textbf{"language\_plan":}@*) "1. Left arm: move above green block, lower, close to grasp, raise. Move to center at [0, -0.13], set orientation for placing, lower to z=0.88649, open to place as bottom, raise and move back to origin high. 2. Right arm: move above red block, lower, close to grasp, raise. Move above center, lower to z=0.93649 to place on top of the green block, open, raise and move back to origin high. Both arms will maintain safe z>=1.05 when moving across the table to avoid collision.",
    (*@\textbf{"executable\_plan":}@*) [
        "[-0.174, 0.029, 0.98, 0.56668, -0.42293, 0.56668, 0.42293, 1.0, 0.3505, -0.2523, 1.08, 0.70711, 0.00003, -0.00003, 0.70711, 1.0]",
        "[-0.174, 0.029, 0.90, 0.56668, -0.42293, 0.56668, 0.42293, 1.0, 0.3505, -0.2523, 1.08, 0.70711, 0.00003, -0.00003, 0.70711, 1.0]",
        "[-0.174, 0.029, 0.90, 0.56668, -0.42293, 0.56668, 0.42293, 0.0, 0.3505, -0.2523, 1.08, 0.70711, 0.00003, -0.00003, 0.70711, 1.0]",
        "[0.0, -0.13, 1.05, 0.5, -0.5, 0.5, 0.5, 0.0, 0.3505, -0.2523, 1.08, 0.70711, 0.00003, -0.00003, 0.70711, 1.0]",
        "[0.0, -0.13, 0.88649, 0.5, -0.5, 0.5, 0.5, 0.0, 0.3505, -0.2523, 1.08, 0.70711, 0.00003, -0.00003, 0.70711, 1.0]",
        "[0.0, -0.13, 0.88649, 0.5, -0.5, 0.5, 0.5, 1.0, 0.3505, -0.2523, 1.08, 0.70711, 0.00003, -0.00003, 0.70711, 1.0]",
        "[-0.3495, -0.2523, 1.08, 0.70711, -0.00001, 0.00001, 0.70711, 1.0, 0.3505, -0.2523, 1.10, 0.70711, 0.00003, -0.00003, 0.70711, 1.0]",
        "[0.002, -0.124, 0.98, 0.49911, -0.50089, 0.49911, 0.50089, 1.0, 0.3505, -0.2523, 1.10, 0.70711, 0.00003, -0.00003, 0.70711, 1.0]",
        "[0.002, -0.124, 0.90, 0.49911, -0.50089, 0.49911, 0.50089, 1.0, 0.3505, -0.2523, 1.10, 0.70711, 0.00003, -0.00003, 0.70711, 1.0]",
        "[0.002, -0.124, 0.90, 0.49911, -0.50089, 0.49911, 0.50089, 1.0, 0.3505, -0.2523, 1.10, 0.70711, 0.00003, -0.00003, 0.70711, 0.0]",
        "[0.0, -0.13, 1.05, 0.5, -0.5, 0.5, 0.5, 1.0, 0.0, -0.13, 1.05, 0.5, -0.5, 0.5, 0.5, 0.0]",
        "[0.0, -0.13, 0.93649, 0.5, -0.5, 0.5, 0.5, 1.0, 0.0, -0.13, 0.93649, 0.5, -0.5, 0.5, 0.5, 0.0]",
        "[0.0, -0.13, 0.93649, 0.5, -0.5, 0.5, 0.5, 1.0, 0.0, -0.13, 0.93649, 0.5, -0.5, 0.5, 0.5, 1.0]",
        "[-0.3495, -0.2523, 1.08, 0.70711, -0.00001, 0.00001, 0.70711, 1.0, 0.3505, -0.2523, 1.08, 0.70711, -0.00001, 0.00001, 0.70711, 1.0]"
    ]
}
\end{lstlisting}

%% file: appendix/prompts/spatial_prompt.tex
\begin{lstlisting}[style=promptstyle, caption={Prompt Template for Dual-Arm Spatial Reasoning}]
You are a dual-arm robot manipulation assistant. You are designed to finish dual-arm manipulation task. However, now you just need to analyze the given observation image and decide which arm (left or right) should perform the grasping action.

The observation image will contain two robotic arms and some objects on a table. You need to analyze the observation image and determine which robotic arm should be used to grasp some given objects.

Your output should be json format and clearly indicate:

VISUAL_STATE_DESCRIPTION (describe what you see) and RESULTS. RESULTS should be a array. Every item in array is a json object which contain OBJECT and its USE_ARM.

!!! Json format is strict. When generating content for JSON strings, avoid using any contractions or abbreviated forms (like 's, 're, 've, 'll, 'd, n't) that use apostrophes. Instead, write out full forms (is, are, have, will, would, not) to prevent parsing errors in JSON. Please do not output any other thing more than the above-mentioned JSON, do not include json code fences.

!!! And do not output any quotation marks in JSON text content (in visual_state_description no quote), like red_block. You can use RED_BLOCK or (red_block) instead. Else json object cannot be loaded successfully. Pay attention to make your output JSON identifiable.

# This is a output example for this task ((LEFT OR RIGHT) means you should choose one):
{
    visual_state_description: (INDICATE WHAT YOU SEE IN PICTURE AND DESCRIBE THE APPROXIMATE LOCATION OF THE BLOCKS, LIKE ON THE LEFT PART OR RIGHT PART),
    results: [
        {
            object: red_block,
            use_arm: (LEFT OR RIGHT)
        },
        {
            object: green_block,
            use_arm: (LEFT OR RIGHT)
        },
        {
            object: blue_block,
            use_arm: (LEFT OR RIGHT)
        }
    ]
}
\end{lstlisting}

%% file: appendix/prompts/highlevel_prompt.tex
\begin{lstlisting}[style=promptstyle, caption={Prompt Template for High-Level Action Planning: Blocks\_ranking\_rgb}]
The task name is blocks_ranking_rgb, the description and the goal of the task is as follows:
<Place> the red block, green block, and blue block <in the order> of red, green, and blue from left to right, <placing in a row>.
You are a robot that can interact with objects in a simulated environment. Your task is to follow the instructions provided to you and complete the task successfully. You will be given a series of actions to perform, and you must execute them in the correct order to achieve the goal. You will also be provided with a set of observations that will help you understand the current state of the environment and the objects within it. Use these observations to inform your actions and make decisions about how to proceed. Your ultimate goal is to successfully complete the task by following the instructions and achieving the desired outcome. Remember to pay attention to the details of the task and the observations provided, as they will guide you in making the right choices. 

!!! You are a dual-arm robot. Please note: When making a plan, do not let your two hands collide with each other. When you do not need to use one of your hands, please let it return to its original position to avoid a collision that may cause the plan to fail.

!!!!! Assistant info (Very Important to provide some key info): This is a two arm task. You must use both arms(left and right) to finish the task. 
The main steps of this task: 1. grasp blocks with nearest arm. 2. raise your hand in case of collision with other objects 3. place one block to ideal position, raise the arm first and then make arm back to origin 4. repeat steps above and place another blocks to ideal positions.  
 In parameter (actor), you must output red_block, green_block or blue_block to represent blocks in different colors. 
!!!! Now [-0.08868774943139059, -0.19668158161348412, 0.737351797170751, 0, 1, 0, 0] is the target_pose of red_block, [0.0016719173640308265, -0.19668158161348412, 0.737351797170751, 0, 1, 0, 0] is the target_pose of green_block and [0.08271717895684488, -0.19668158161348412, 0.737351797170751, 0, 1, 0, 0] is the target_pose of blue_block. You can use them *directly* in the target_pose of place_actor() function. It's 7-dim. You can also try to use the first 3 dimensions for the target_pose. 
You must raise the robot arm a certain distance to move the object or execute place command, otherwise it may hit something if it moves directly close to the table surface. But you should not raise the arm too high, otherwise the object may fall off or the raise action will fail. 
 You can use the action 'back_to_origin' to return the arm to the origin position. 
 !!! Besides, if you don't need arm right now, *please make arm back to origin* in dual arm tasks. Else it may block actions of another arm. 
 !!!NOTE: In parameters of PLACE_ACTOR() function, you must specify the functional_point_id = 0 and constrain='align', else the place action will be failed!!! And you are recommended to set pre_grasp_dis=0.09 and grasp_dis=0.03 to make it easier to success. When you make arm back to origin, you should raise the arm first to avoid collision with other objects on the table.

The images in every observation are the current state of the environment, involving the image from head_camera, front_camera, third_view_camera, and the example image of the task to help you make decisions.
The output json format should be {'visual_state_description':str, 'reasoning_and_reflection':str, 'language_plan':str, 'executable_plan': str}
The fields in above JSON follows the purpose below:
1. visual_state_description is for description of current state from the visual image, and you should describe the gripper state by your eyes. 
2. reasoning_and_reflection is for summarizing the history of interactions and any available environmental feedback. Additionally, provide reasoning as to why the last action or plan failed and did not finish the task, 
3. language_plan is for describing the following action to achieve the user instruction.
4. executable_plan is an json array that contains the next action to achieve the user instruction. Every item of the json array is a json object that contains the action id (2.2 to 2.9) and the according action name. Additionally, you must provide the parameters for the action function. The instruction for the action function and parameters is as follows:

Action list is following, action id from 2.2 to 2.9:
{
  "2.2": {
    "name": "grasp_actor",
    "description": "Pick up a specified object using the selected arm.",
    "parameters": {
      "actor": "The object to grasp.",
      "arm_tag": "Which arm to use.",
      "pre_grasp_dis": "Pre-grasp distance (default 0.1 meters), the arm will move to this position first.",
      "grasp_dis": "Grasping distance (default 0 meters), the arm moves from the pre-grasp position to this position and then closes the gripper.",
      "gripper_pos": "Gripper closing position (default 0, fully closed).",
      "contact_point_id": "Optional list of contact point IDs; if not provided, the best grasping point is selected automatically."
    },
    "returns": "tuple[ArmTag, list[Action]]",
    "example": "self.move(self.grasp_actor(self.cup, arm_tag=arm_tag, pre_grasp_dis=0.1, contact_point_id=[0, 2][int(arm_tag=='left')]))"
  },
  "2.3": {
    "name": "place_actor",
    "description": "Places a currently held object at a specified target pose.",
    "parameters": {
      "actor": "The currently held object.",
      "arm_tag": "The arm holding the object.",
      "target_pose": "Target position/orientation, length 3 or 7 (xyz + optional quaternion).",
      "functional_point_id": "Optional ID of the functional point; aligns this point to the target if provided. Please don't provide. It will be ignored. Use target_pose to finish this action task.",
      "pre_dis": "Pre-place distance (default 0.1 meters).",
      "dis": "Final placement distance (default 0.02 meters).",
      "is_open": "Whether to open the gripper after placing (default True).",
      "kwargs": {
        "constrain": "Alignment strategy: 'free', 'align', or 'auto' (default).",
        "align_axis": "Vectors in world coordinates to align with.",
        "actor_axis": "Second object axis used for alignment (default [1, 0, 0]).",
        "actor_axis_type": "Whether actor_axis is relative to 'actor' or 'world' (default 'actor').",
        "pre_dis_axis": "Direction of pre-placement offset: 'grasp', 'fp', or custom vector."
      }
    },
    "returns": "tuple[ArmTag, list[Action]]",
    "example": "self.move(self.place_actor(actor=self.current_actor, target_pose=target_pose, arm_tag=arm_tag, functional_point_id=0, pre_dis=0.1, dis=0.02, pre_dis_axis='fp'))"
  },
  "2.4": {
    "name": "move_by_displacement",
    "description": "Moves the end-effector of the specified arm along relative directions and sets its orientation.",
    "parameters": {
      "arm_tag": "The arm to control.",
      "x": "Displacement along x-axis (meters).",
      "y": "Displacement along y-axis (meters).",
      "z": "Displacement along z-axis (meters).",
      "quat": "Optional quaternion specifying the target orientation.",
      "move_axis": "'world' or 'arm'; defines coordinate system (default 'world')."
    },
    "returns": "tuple[ArmTag, list[Action]]",
    "example": "self.move(self.move_by_displacement(arm_tag=arm_tag, z=0.07, move_axis='world'))"
  },
  "2.5": {
    "name": "move_to_pose",
    "description": "Moves the end-effector of the specified arm to a specific absolute pose.",
    "parameters": {
      "arm_tag": "The arm to control.",
      "target_pose": "Absolute pose (xyz + optional quaternion)."
    },
    "returns": "tuple[ArmTag, list[Action]]",
    "example": "self.move(self.move_to_pose(arm_tag=arm_tag, target_pose=target_pose))"
  },
  "2.6": {
    "name": "close_gripper",
    "description": "Closes the gripper of the specified arm.",
    "parameters": {
      "arm_tag": "Which arm's gripper to close.",
      "pos": "Gripper position (0 = fully closed)."
    },
    "returns": "tuple[ArmTag, list[Action]]",
    "example": "self.move(self.close_gripper(arm_tag=arm_tag))"
  },
  "2.7": {
    "name": "open_gripper",
    "description": "Opens the gripper of the specified arm.",
    "parameters": {
      "arm_tag": "Which arm's gripper to open.",
      "pos": "Gripper position (1 = fully open)."
    },
    "returns": "tuple[ArmTag, list[Action]]",
    "example": "self.move(self.open_gripper(arm_tag=arm_tag))"
  },
  "2.8": {
    "name": "back_to_origin",
    "description": "Returns the specified arm to its predefined initial position.",
    "parameters": {
      "arm_tag": "The arm to return to origin."
    },
    "returns": "tuple[ArmTag, list[Action]]",
    "example": "self.move(self.back_to_origin(arm_tag=ArmTag('right')))"
  },
  "2.9": {
    "name": "get_arm_pose",
    "description": "Gets the current pose of the end-effector of the specified arm.",
    "parameters": {
      "arm_tag": "Which arm to query."
    },
    "returns": "list[float]",
    "example": "pose = self.get_arm_pose(ArmTag('left'))"
  }
}

!!! When generating content for JSON strings, avoid using any contractions or abbreviated forms (like 's, 're, 've, 'll, 'd, n't) that use apostrophes. Instead, write out full forms (is, are, have, will, would, not) to prevent parsing errors in JSON. Please do not output any other thing more than the above-mentioned JSON, do not include ```json and ```!!!.
!!! And do not output any ""(quotation marks) in JSON content like "can_right", you can use can_right or (can_right) instead. Else json object can't be loaded successfully and the action will be empty. 
# This is a proper output template example: 

{OUTPUT_EXAMPLE}

\end{lstlisting}

Depending on the task state, the prompt is constructed conditionally.
Specifically, when prior execution steps are available, feedback from the three most recent actions is appended to the prompt; otherwise, this component is omitted.

\begin{lstlisting}[style=promptstyle, caption={3-steps feedback prompt template for High-Level Action Planning: Blocks\_ranking\_rgb}]

The 3-steps action history:
Step 1, actionList [{'action_id': '2.2', 'action_name': 'grasp_actor', 'parameters': {'actor': 'red_block', 'arm_tag': 'right', 'pre_grasp_dis': 0.09, 'grasp_dis': 0.03}}, {'action_id': '2.4', 'action_name': 'move_by_displacement', 'parameters': {'arm_tag': 'right', 'z': 0.1, 'move_axis': 'world'}}, {'action_id': '2.3', 'action_name': 'place_actor', 'parameters': {'actor': 'red_block', 'arm_tag': 'right', 'target_pose': [-0.08000283043448461, -0.14068537547170237, 0.7190565302370682, 0, 1, 0, 0], 'functional_point_id': 0, 'constrain': 'align', 'pre_dis': 0.09, 'dis': 0.03}}, {'action_id': '2.4', 'action_name': 'move_by_displacement', 'parameters': {'arm_tag': 'right', 'z': 0.1, 'move_axis': 'world'}}, {'action_id': '2.8', 'action_name': 'back_to_origin', 'parameters': {'arm_tag': 'right'}}], action_feedback:Action succeeded.
Action succeeded.
Action succeeded.
Action succeeded.
Action succeeded.

Step 2, actionList [{'action_id': '2.2', 'action_name': 'grasp_actor', 'parameters': {'actor': 'green_block', 'arm_tag': 'right', 'pre_grasp_dis': 0.09, 'grasp_dis': 0.03}}, {'action_id': '2.4', 'action_name': 'move_by_displacement', 'parameters': {'arm_tag': 'right', 'z': 0.1, 'move_axis': 'world'}}, {'action_id': '2.3', 'action_name': 'place_actor', 'parameters': {'actor': 'green_block', 'arm_tag': 'right', 'target_pose': [-0.006194052012989204, -0.14068537547170237, 0.7190565302370682, 0, 1, 0, 0], 'functional_point_id': 0, 'constrain': 'align', 'pre_dis': 0.09, 'dis': 0.03}}, {'action_id': '2.4', 'action_name': 'move_by_displacement', 'parameters': {'arm_tag': 'right', 'z': 0.1, 'move_axis': 'world'}}, {'action_id': '2.8', 'action_name': 'back_to_origin', 'parameters': {'arm_tag': 'right'}}], action_feedback:Action failed: Action Failed: target green_block is too far, right arm can not finish this 'grasp' action! Please use another arm!

Step 3, actionList [{'action_id': '2.2', 'action_name': 'grasp_actor', 'parameters': {'actor': 'green_block', 'arm_tag': 'left', 'pre_grasp_dis': 0.09, 'grasp_dis': 0.03}}, {'action_id': '2.4', 'action_name': 'move_by_displacement', 'parameters': {'arm_tag': 'left', 'z': 0.1, 'move_axis': 'world'}}, {'action_id': '2.3', 'action_name': 'place_actor', 'parameters': {'actor': 'green_block', 'arm_tag': 'left', 'target_pose': [-0.006194052012989204, 0.14068537547170237, 0.7190565302370682, 0, 1, 0, 0], 'functional_point_id': 0, 'constrain': 'align', 'pre_dis': 0.09, 'dis': 0.03}}, {'action_id': '2.4', 'action_name': 'move_by_displacement', 'parameters': {'arm_tag': 'left', 'z': 0.1, 'move_axis': 'world'}}, {'action_id': '2.8', 'action_name': 'back_to_origin', 'parameters': {'arm_tag': 'left'}}, {'action_id': '2.2', 'action_name': 'grasp_actor', 'parameters': {'actor': 'blue_block', 'arm_tag': 'right', 'pre_grasp_dis': 0.09, 'grasp_dis': 0.03}}, {'action_id': '2.4', 'action_name': 'move_by_displacement', 'parameters': {'arm_tag': 'right', 'z': 0.1, 'move_axis': 'world'}}, {'action_id': '2.3', 'action_name': 'place_actor', 'parameters': {'actor': 'blue_block', 'arm_tag': 'right', 'target_pose': [0.08393771486567518, 0.14068537547170237, 0.7190565302370682, 0, 1, 0, 0], 'functional_point_id': 0, 'constrain': 'align', 'pre_dis': 0.09, 'dis': 0.03}}, {'action_id': '2.4', 'action_name': 'move_by_displacement', 'parameters': {'arm_tag': 'right', 'z': 0.1, 'move_axis': 'world'}}, {'action_id': '2.8', 'action_name': 'back_to_origin', 'parameters': {'arm_tag': 'right'}}], action_feedback:Action succeeded.
Action succeeded.
Action succeeded.
Action succeeded.
Action succeeded.
Action succeeded.
Action succeeded.
Action succeeded.
Action succeeded.
Action succeeded.

 Considering the above interaction history and the current image state, to achieve the human instruction: '<Place> the red block, green block, and blue block <in the order> of red, green, and blue from left to right, <placing in a row>.', you are supposed to output in json. You need to describe current visual state from the image, summarize interaction history and environment feedback and reason why the last action or plan failed and did not finish the task, output your new plan to achieve the goal from current state. At the end, output the actions.
\end{lstlisting}

%% file: appendix/prompts/lowlevel_prompt.tex
\begin{lstlisting}[style=promptstyle, caption={Prompt Template for Low-Level End-Effector Controlling: Stacks\_blocks\_two}]

You are a robot manipulation assistant. Your task is to follow the instruction and move objects using **End-Effector Pose Control mode** for a dual-arm robot.
The action format is:
```
[left_end_effector_pose (xyz + quaternion) + left_gripper + right_end_effector_pose (xyz + quaternion) + right_gripper]
```
!!!NOTE: Please control the float number in output to be within 5 digits.
### Parameter Explanation:
1. **xyz**
   The position of the end-effector in the world coordinate system (unit: meters). Example: `[0.2, 0.3, 0.1]`.
2. **quaternion (qx, qy, qz, qw)**
   The orientation of the end-effector, represented as a quaternion. The first three values `(qx, qy, qz)` describe the rotation axis (a unit vector), and the last value `qw` is the cosine of half the rotation angle:
   [
   q = [\sin(\theta/2) \cdot u_x,; \sin(\theta/2) \cdot u_y,; \sin(\theta/2) \cdot u_z,; \cos(\theta/2)]
   ]
**Example**: A 90(*@\textdegree@*) rotation around the z-axis is:
[
[0.0, 0.0, \sin(90(*@\textdegree@*)/2), \cos(90(*@\textdegree@*)/2)] = [0.0, 0.0, 0.7071, 0.7071]
]
We use a **right-handed coordinate system**: if you point your right thumb along the positive axis, the curl of your fingers indicates the positive rotation direction.
Thus, looking **from the positive z-axis downward**, a +90(*@\textdegree@*) rotation is **counterclockwise**.

3. **Coordinate System Convention**
   In the environment visualization:
* Right side = positive x-axis
* Into the screen = positive y-axis
* Upward = positive z-axis
4. **gripper (open/close state)**
   Range `[0, 1]`:

* `0` = fully closed
* `1` = fully open
* e.g., `0.5` = half open.
---
### Action Example:
Scenario: An apple is located at `[0.2, 0.3, 0.0]`.
* **Action 1**
```
[0.2, 0.3, 0.15, 0.5, -0.5, 0.5, 0.5, 1.0, 0.5, 0.3, 0.1, 0.0, 0.5, -0.5, 0.5, 0.5]
```
Explanation: Move the left hand above the apple.
* **Action 2**
```
[0.2, 0.3, 0.08, 0.5, -0.5, 0.5, 0.5, 1.0, 0.5, 0.3, 0.1, 0.0, 0.5, -0.5, 0.5, 0.5]
```
Explanation: Lower the left hand to grasp the apple.

* **Action 3**
```
[0.2, 0.3, 0.08, 0.5, -0.5, 0.5, 0.5, 0, 0.5, 0.3, 0.1, 0.0, 0.5, -0.5, 0.5, 0.5]
```
Explanation: Close the left gripper to pick up the apple.
---
!!! When generating content for JSON strings, avoid using any contractions or abbreviated forms (like 's, 're, 've, 'll, 'd, n't) that use apostrophes. Instead, write out full forms (is, are, have, will, would, not) to prevent parsing errors in JSON. Please do not output any other thing more than the above-mentioned JSON, do not include ```json and ```!!!.
!!! And do not output any ""(quotation marks) in JSON content like "can_right", you can use can_right or (can_right) instead. Else json object can't be loaded successfully and the action will be empty. 
!!! you must control your output length within 3000 tokens. Else the output will be cut off and the action will be empty. And output nothing else except the json format below.
!!! The max step you can manipulate is limited, so try your best to use dual arm at the same time to finish the task. And try to output more than an action in every output.
# This is a proper output template example: 
{
    "visual_state_description": "(INDICATE WHAT YOU SEE IN PICTURE.)",
    "reasoning_and_reflection": "(DESCRIBE YOUR THOUGHT ACCORDING TO OBSERVATION AND ASSISTANT INFO. DETERMINE WHICH STEP THE TASK HAS REACHED. THINKING THE IDEA TO FINISH THE TASK. DESCRIBE ARM TO USE TO PLACE EVERY OBJECT ACCORDING TO YOUR OBSERVATION)",
    "language_plan": "(DESCRIBE YOUR PLAN TO FINISH THE TASK WITH LANGUAGE)",
    "executable_plan": [
        "[-0.25, -0.09, 0.98, 0.5, -0.5, 0.5, 0.5, 1.0, 0.22, -0.083, 0.98, 0.5, -0.5, 0.5, 0.5, 1.0]",
        "[-0.25, -0.09, 0.9, 0.5, -0.5, 0.5, 0.5, 1.0, 0.22, -0.07, 0.98, 0.5, -0.5, 0.5, 0.5, 1.0]",
        "[-0.25, -0.09, 0.9, 0.5, -0.5, 0.5, 0.5, 0.0, 0.22, -0.07, 0.9, 0.5, -0.5, 0.5, 0.5, 1.0]",
        "[-0.25, -0.09, 1.0, 0.5, -0.5, 0.5, 0.5, 0.0, 0.22, -0.07, 0.9, 0.5, -0.5, 0.5, 0.5, 0.0]"
    ]
}

!!!!! Assistant info (Very Important to provide some key info): This is a two arm task. You must use both arms(left and right) to finish the task. 
The main steps of this task: 1.move grippers above the object(use nearest arm to grasp object on the same side) 2.lower the gripper to prepare to grasp the objects 3.close the gripper to grasp the object 4.raise the arm and keep gripper close 5.move arm above the target position and keep gripper close 6.release the gripper and place object to proper positions 7.raise arm and move arm back to origin 8.choose the proper arm to grasp another object next and follow steps above
Now the left arm pose is [-3.4950e-01 -2.5230e-01  9.4049e-01  7.0711e-01 -1.0000e-05  1.0000e-05
  7.0711e-01] and the right arm pose is [ 3.5050e-01 -2.5230e-01  9.4049e-01  7.0711e-01 -1.0000e-05  1.0000e-05
  7.0711e-01].
Now the positions of block1 is [-0.19881 -0.07057  0.73949] and of block2 is [-0.17402  0.02905  0.73949]. You should judge the position of two blocks and use proper arm to grasp. When you try to grasp block1, you should set q or quat=[ 0.64743 -0.2843   0.64743  0.2843 ]. When you try to grasp block2, you should set quat=[ 0.56669 -0.42293  0.56669  0.42293]. However, you are recommended to set quat = [0.5, -0.5, 0.5, 0.5] when you place block to center.
The center target position(x,y) is [0, -0.13]. Your target is to stack two blocks. When you place the bottom block, its ideal pos of z is 0.7244852714832615. But this isn't the gripper pos which you manipulate or control. There's a distance difference between the gripper's center position and its lowest point, meaning the gripper's height is approximately 0.162m. So when you place the bottom block, you should add 0.162 in z, which will be the final proper z-pos. And when you place the second block on top, its ideal pos of z is 0.7744852714832615(5 cm higher than bottom block). And you should also add 0.162 in z when you control gripper pos. If you don't add 0.162m in z of pos, there will be some errors about arm's action and your instruction will fail.
NOTE: The initial arm pose of left arm is [-3.4950e-01 -2.5230e-01  9.4049e-01  7.0711e-01 -1.0000e-05  1.0000e-05
  7.0711e-01] and of right arm is [ 3.5050e-01 -2.5230e-01  9.4049e-01  7.0711e-01 -1.0000e-05  1.0000e-05
  7.0711e-01]. You can use these poses to make arm back to origin. The original gripper state is open or 1. This will be important in this task!!! Raise arm high enough first(z>=1.05m) and then make arm back to origin when you don't need it, otherwise it will block the actions of another arm!!!!! So raise arm high enough(1.05 or 1.1m) and then make arm back to origin if you don't need it now!!!
!!!!!!NOTE: There's a distance difference between the gripper's center position and its lowest point, meaning the gripper's height is approximately 0.162m. You can't lower the gripper too low. For example, if the object's height (z) is 0.738m, and you should output a pose with a lowest point of 0.9m, you need to add 0.162 and set z=0.9 to prevent the gripper from hitting the table and causing damage. In other words, when you output a pose with a z value of 1m, the lowest point is 0.838m,. This involves some calculations. In the action, you must calculate and output the final result in advance, rather than outputting the expression for me to calculate. Note that your output is directly submitted to the environment for interaction, so make sure your output conforms to the format requirements! If you output like 0.738+0.162, this will be illegal.
When you manipulate the arm to place center, you are recommended to set q or orientation(last 4-dim)=[0.5,-0.5,0.5,0.5] in most cases. This will point the grippers downwards and can open gripper left-right for easier placing. When you manipulate the arm to grasp blocks, you should set q or quat according to the assistant info above.
Since there is a certain error in the End-Effector Pose Control mode (maybe several millimeters), please do not trust the information provided in the assistant info too much. If the observation shows obvious deviation, please move it in a more appropriate direction based on the position provided in the assistant info (for example, add 0.1 (m) to the positive direction of the y-axis).
Don't output too many actions in an output. The position of object will change when your former actions are executed. Make plan based on the latest observation.
NOTE: To avoid a fight between two robotic arms, please consider whether the two will collide when operating the robotic arms. If there is a possibility of collision, operate one robotic arm first, move it away after the operation, and then operate the other robotic arm.

{OUTPUT_EXAMPLE}

\end{lstlisting}

Depending on the task state, the prompt is constructed conditionally.
Specifically, when prior execution steps are available, feedback from the three most recent actions is appended to the prompt; otherwise, this component is omitted.

\begin{lstlisting}[style=promptstyle, caption={3-steps feedback prompt template for Low-Level End-Effector Controlling: Stack\_blocks\_two}]
The 3-steps action history:
Step 0, actionList ['[-0.19881, -0.07057, 0.94049, 0.64743, -0.2843, 0.64743, 0.2843, 1.0, 0.35050, -0.25230, 0.94049, 0.70711, -0.00001, 0.00001, 0.70711, 1.0]', '[-0.19881, -0.07057, 0.73949, 0.64743, -0.2843, 0.64743, 0.2843, 1.0, 0.35050, -0.25230, 0.94049, 0.70711, -0.00001, 0.00001, 0.70711, 1.0]', '[-0.19881, -0.07057, 0.73949, 0.64743, -0.2843, 0.64743, 0.2843, 0.0, 0.35050, -0.25230, 0.94049, 0.70711, -0.00001, 0.00001, 0.70711, 1.0]', '[-0.19881, -0.07057, 1.05, 0.64743, -0.2843, 0.64743, 0.2843, 0.0, 0.35050, -0.25230, 0.94049, 0.70711, -0.00001, 0.00001, 0.70711, 1.0]', '[0.0, -0.13, 1.05, 0.5, -0.5, 0.5, 0.5, 0.0, 0.35050, -0.25230, 0.94049, 0.70711, -0.00001, 0.00001, 0.70711, 1.0]', '[0.0, -0.13, 0.88649, 0.5, -0.5, 0.5, 0.5, 0.0, 0.35050, -0.25230, 0.94049, 0.70711, -0.00001, 0.00001, 0.70711, 1.0]', '[0.0, -0.13, 0.88649, 0.5, -0.5, 0.5, 0.5, 1.0, 0.35050, -0.25230, 0.94049, 0.70711, -0.00001, 0.00001, 0.70711, 1.0]', '[-0.34950, -0.25230, 1.05, 0.70711, -0.00001, 0.00001, 0.70711, 1.0, 0.35050, -0.25230, 0.94049, 0.70711, -0.00001, 0.00001, 0.70711, 1.0]', '[-0.34950, -0.25230, 1.05, 0.70711, -0.00001, 0.00001, 0.70711, 1.0, -0.17402, 0.02905, 0.94049, 0.56669, -0.42293, 0.56669, 0.42293, 1.0]'], action_feedback:Action succeeded.
Action succeeded.
Action succeeded.
Action succeeded.
Action succeeded.
Action succeeded.
Action succeeded.
Action succeeded.
Action succeeded.

Step 1, actionList ['[ -0.34950, -0.25230, 1.05, 0.70711, -0.00001, 0.00001, 0.70711, 1.0, -0.00823, -0.14638, 1.05, 0.47029, -0.52805, 0.47029, 0.52805, 1.0]', '[ -0.34950, -0.25230, 1.05, 0.70711, -0.00001, 0.00001, 0.70711, 1.0, -0.00823, -0.14638, 0.90149, 0.47029, -0.52805, 0.47029, 0.52805, 1.0]', '[ -0.34950, -0.25230, 1.05, 0.70711, -0.00001, 0.00001, 0.70711, 1.0, -0.00823, -0.14638, 0.90149, 0.47029, -0.52805, 0.47029, 0.52805, 0.0]', '[ -0.34950, -0.25230, 1.05, 0.70711, -0.00001, 0.00001, 0.70711, 1.0, -0.00823, -0.14638, 1.05, 0.47029, -0.52805, 0.47029, 0.52805, 0.0]', '[ -0.34950, -0.25230, 1.05, 0.70711, -0.00001, 0.00001, 0.70711, 1.0, 0.0, -0.13, 1.05, 0.5, -0.5, 0.5, 0.5, 0.0]', '[ -0.34950, -0.25230, 1.05, 0.70711, -0.00001, 0.00001, 0.70711, 1.0, 0.0, -0.13, 0.88649, 0.5, -0.5, 0.5, 0.5, 0.0]', '[ -0.34950, -0.25230, 1.05, 0.70711, -0.00001, 0.00001, 0.70711, 1.0, 0.0, -0.13, 0.88649, 0.5, -0.5, 0.5, 0.5, 1.0]', '[ -0.34950, -0.25230, 1.05, 0.70711, -0.00001, 0.00001, 0.70711, 1.0, 0.35050, -0.25230, 1.05, 0.70711, -0.00001, 0.00001, 0.70711, 1.0]'], action_feedback:Action succeeded.
Action succeeded.
Action succeeded.
Action succeeded.
Action succeeded.
Action succeeded.
Action succeeded.
Action succeeded.

Step 2, actionList ['[-0.34950, -0.25230, 1.05, 0.70711, -0.00001, 0.00001, 0.70711, 1.0, 0.01411, -0.1228, 1.05, 0.52283, -0.47608, 0.52283, 0.47608, 1.0]', '[-0.34950, -0.25230, 1.05, 0.70711, -0.00001, 0.00001, 0.70711, 1.0, 0.01411, -0.1228, 0.90149, 0.52283, -0.47608, 0.52283, 0.47608, 1.0]', '[-0.34950, -0.25230, 1.05, 0.70711, -0.00001, 0.00001, 0.70711, 1.0, 0.01411, -0.1228, 0.90149, 0.52283, -0.47608, 0.52283, 0.47608, 0.0]', '[-0.34950, -0.25230, 1.05, 0.70711, -0.00001, 0.00001, 0.70711, 1.0, 0.01411, -0.1228, 1.05, 0.52283, -0.47608, 0.52283, 0.47608, 0.0]', '[-0.34950, -0.25230, 1.05, 0.70711, -0.00001, 0.00001, 0.70711, 1.0, 0.0, -0.13, 1.05, 0.5, -0.5, 0.5, 0.5, 0.0]', '[-0.34950, -0.25230, 1.05, 0.70711, -0.00001, 0.00001, 0.70711, 1.0, 0.0, -0.13, 0.88649, 0.5, -0.5, 0.5, 0.5, 0.0]', '[-0.34950, -0.25230, 1.05, 0.70711, -0.00001, 0.00001, 0.70711, 1.0, 0.0, -0.13, 0.88649, 0.5, -0.5, 0.5, 0.5, 1.0]', '[-0.34950, -0.25230, 1.05, 0.70711, -0.00001, 0.00001, 0.70711, 1.0, 0.35050, -0.25230, 1.05, 0.70711, -0.00001, 0.00001, 0.70711, 1.0]'], action_feedback:Action succeeded.
Action succeeded.
Action succeeded.
Action succeeded.
Action succeeded.
Action succeeded.
Action succeeded.
Action succeeded.

 Considering the above interaction history and the current image state, to achieve the human instruction: 'there are two blocks on the table, the color of the blocks is <red, green>, <move the blocks to the center of the table>, and stack the blocks.', you are supposed to output in json. You need to describe current visual state from the image, summarize interaction history and environment feedback and reason why the last action or plan failed and did not finish the task, output your new plan to achieve the goal from current state. At the end, output the actions.

\end{lstlisting}